\documentclass[smallcondensed]{svjour3}
\usepackage{url}
\usepackage{amsmath,amssymb,amsfonts}
\usepackage{algorithmic}
\usepackage{graphicx,multirow}
\usepackage{textcomp}
\usepackage{float}
\usepackage{xcolor,url}
\graphicspath{{figs/}}

\begin{document}
\title{Explaining Classifiers by Constructing Familiar Concepts\footnote{This is an extended journal version of the conference publication ``Explaining Neural Networks by Decoding Layer Activations'', accepted at the Intelligent Data Analysis Symposium(IDA), 2021} }

\author{Johannes Schneider \and Michalis Vlachos }
\institute{J. Schneider \at 
Institute of Information Systems, University of Liechtenstein, Liechtenstein \\ \email{johannes.schneider@uni.li}
\and M. Vlachos \at 
Department of Information Systems, HEC, University of Lausanne, Switzerland \\ 
\email{michalis.vlachos@unil.ch} }

%

\date{}
\maketitle

\begin{abstract} 
Interpreting a large number of neurons in deep learning is difficult. Our proposed `CLAssifier-DECoder' architecture (\emph{ClaDec}) facilitates the understanding of the output of an arbitrary layer of neurons or subsets thereof. It uses a decoder that transforms the incomprehensible representation of the given neurons to a representation that is more similar to the domain a human is familiar with.  In an image recognition problem, one can recognize what information (or concepts) a layer maintains by contrasting reconstructed images of \emph{ClaDec} with those of a conventional auto-encoder(AE) serving as reference. An extension of \emph{ClaDec}  allows trading comprehensibility and fidelity. We evaluate our approach for image classification using convolutional neural networks. We show that reconstructed visualizations using encodings from a classifier capture more relevant classification information than conventional AEs. This holds although AEs contain more information on the original input. Our user study highlights that even non-experts can identify a diverse set of concepts contained in images that are relevant (or irrelevant) for the classifier. We also compare against saliency based methods that focus on pixel relevance rather than concepts. We show that \emph{ClaDec} tends to highlight more relevant input areas to classification though outcomes depend on classifier architecture. Code is at \url{https://github.com/JohnTailor/ClaDec}
\end{abstract}

\section{Introduction} 
\begin{figure}
  \centering
  \includegraphics[width=0.9\linewidth]{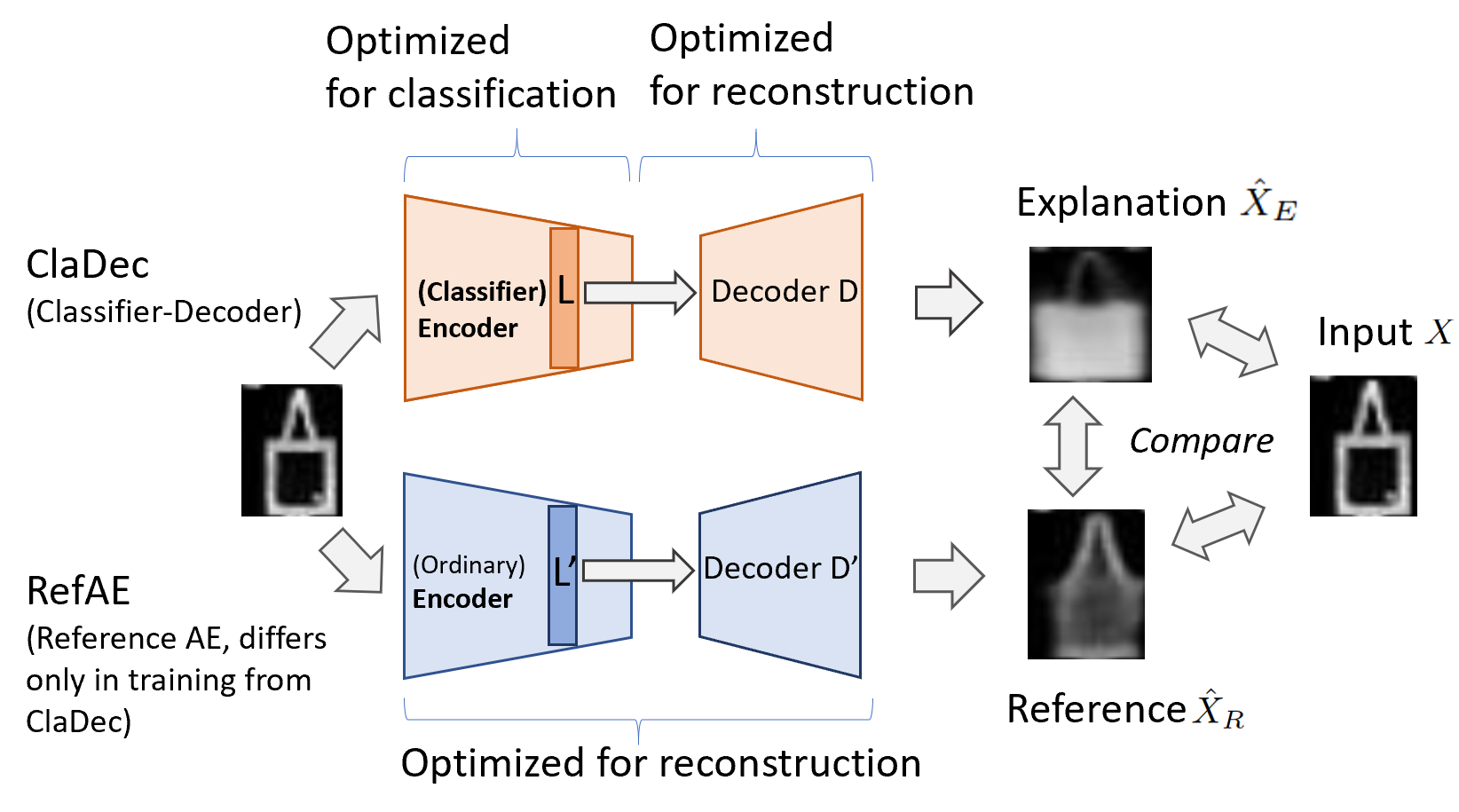}
  \caption{Basic architecture of \emph{ClaDec} and \emph{RefAE} and explanation process} \label{fig:arch}
\end{figure}

Explaining predictive models is important for many reasons, including: a) debugging or improving models, b) fulfilling legal obligations such as the "right to explanation" as crystallized in the European GDPR data privacy law, and c) increasing trust in models. Thus, explaining neural networks has received a lot of attention \cite{adadi2018peeking,schneider2019pers,conf21,schne22d}.
Understanding a neural network is a multi-faceted problem, ranging from understanding single decisions, single neurons and single layers, up to explaining complete models. Often, explainability methods touch on multiple of these aspects. In this work, we are primarily interested in better understanding a decision with respect to a user-defined layer (or a subset thereof) that originates from a complex feature hierarchy, as commonly found in deep learning models. In a layered model, each layer corresponds to a transformed representation of the original input. Thus, the neural network succinctly transforms the input into more useful representations for the task at hand, such as classification. From this point of view, we seek to answer the question: \textit{``Given an input $X$, what does the representation $L(X)$ (or subsets thereof) produced in a layer $L$ tell us about the decision and about the network?''}. 
Our local, model-agnostic post-hoc explanation method requires comparing three images $X, \hat{X}_E$ and $\hat{X}_R$ that together constitute the explanation as illustrated in the right of Figure \ref{fig:arch}. We compare the original input image $X$ with a transformation $\hat{X}_E$ of the latent representation $L(X)$ of the classifier to image space. This allows to visually identify what information is maintained in the layer and what information is discarded. We also compare against a reference $\hat{X}_R$ consisting of an autoencoded image of $X$.  This allows identifying what of the discarded information in $\hat{X}_E$ is a result of inaccuracies of the explanation process.

Technically, we propose a classifier-decoder architecture called \emph{ClaDec}. It uses a decoder to transform the representation $L(X)$ produced by a layer $L$ of the classifier, with the goal to explain the layer representation via a humanly understandable representation, i.e., one that is similar on the input domain. The layer in question provides the ``code''  that is fed into a decoder. The motivation for using an AE is that AEs are good at (re)constructing high-dimensional data from a low-dimensional representation. Intuitively, a classifier to be explained should encode aspects relevant to the classification faithfully and ignore input information that does not impact decisions. Therefore, using a decoder can lead to accurate reconstruction of parts and attributes of the input that are essential for classification. In contrast, inputs with little or no influence on the classification will be reconstructed at lower fidelity. Attributes of an input might refer to basic properties such as color, shape, sharpness but also more abstract, higher-level concepts. That is, reconstructions of higher-level constructs might be altered to be more similar to prototypical, average-like instances.

Explanations should fulfill many partially conflicting objectives. We are interested in the trade-off between fidelity (How accurately does the explanation express the model behavior?) and comprehensibility (How easy is it to make sense of the explanation?). While these properties of explanations are well-known, existing methods typically do not accommodate adjusting this trade-off. In contrast, we propose an extension of our base architecture \emph{ClaDec} by adding a classification loss. It allows for balancing between producing reconstructions that are similar to the inputs, i.e., training data that a user is probably more familiar with (easier interpretation), and reconstructions that are strongly influenced by the model to explain (higher fidelity) but may deviate more from what the user knows or has seen.

Our approach relies on an auxiliary model, a decoder, to provide explanations. Similar to other methods that use auxiliary or proxy models, e.g., to synthesize inputs \cite{ngu16} or approximate model behavior \cite{ribeiro2016should}, we face the problem that a poor auxiliary model may negatively impact explanation fidelity. That is, reconstructions produced by AEs (or GANs) might suffer from artifacts. For example, AEs are known to produce images that might appear more blurry than real images. People have noticed that GANs can produce clearer images but they may suffer from other artifacts as shown in \cite{ngu16}.  Neglecting that the explainability method might introduce artifacts can adversely impact understandability and even lead to wrong conclusions on model behavior. When looking at the reconstruction, a person not familiar with such artifacts might not attribute the distortion to the auxiliary model being used, but she might believe that it is due to the model to be explained. At the same time, evaluation of explainability methods has many known open questions \cite{yang19}, the community has not been aware of this one. To avoid any wrongful perceptions regarding artifacts in reconstruction, we suggest comparing outcomes of auxiliary models to a reference architecture.  We employ an auto-encoder \emph{RefAE} with the exact same architecture as \emph{ClaDec} to generate outputs for comparison as shown in Figure \ref{fig:arch}. The encoder of  \emph{RefAE} is not trained for classification, but the \emph{RefAE} model optimizes the reconstruction loss of the original inputs as any conventional AE. Therefore, only the differences visible in the reconstructions of  \emph{RefAE} and \emph{ClaDec} can be attributed to the model to be explained. The proposed comparison to a reference model can also be perceived as a rudimentary sanity check, i.e., if there are no differences then either the explainability method is of little value or the features to be learned of the model to be explained are similar to that of the reference AE. This might be a consequence of a similar objective of the model to explain and the reference model, or the fact that there are universal features suitable for many tasks. We shall elaborate more in our theoretical motivation. We believe that such sanity checks are urgently needed since multiple explanation methods have been scrutinized for failing ``sanity'' checks and simple robustness properties \cite{adebayo2018sanity,kin19,gho19}. For that reason, we also introduce a sanity check that formalizes the idea that explanations should be beneficial for downstream tasks. In our context, we even show that auxiliary classifiers trained on either reconstructions from \emph{RefAE} or \emph{ClaDec} perform better on the latter, although the reference AE leads to reconstructions that are closer to the original inputs. Thus, the reconstructions of \emph{ClaDec} are more amendable for the task to be solved. \\
While our reconstruction-based method conceptually differs strongly from a saliency-based method, we also quantitatively compare against one of the most prominent attribution techniques, i.e., GradCAM. To this end, we occlude relevant areas of inputs. The relevance score of input areas is well-suited to the idea of attribution-based techniques that assume that classification outcome depends on a subset of all input pixels. To get the relevance of an area, the individual scores of input pixels simply have to be summed. In turn, we argue more about concepts and how the classifier perceives them. Thus, our explanations do not directly yield a relevance score per pixel. Still, by introducing a measure based on reconstruction loss between reconstructed images with and without occlusion, we obtain a suitable relevance score allowing us to compare GradCAM and \emph{ClaDec}. Our findings confirm the strengths of \emph{ClaDec} showing that it better allows finding areas of minimum and maximum relevance, although the evaluation metrics are more tailored to GradCAM. Advantages of \emph{ClaDec} are most profound when layers with limited spatial extent covering semantically meaningful concepts are used. Overall, the methods are less of a substitute but should be used together to better understand classifiers.\\
Our experiment with humans confirms that even non-experts can identify meaningful concepts discarded (and maintained) by a classifier. Overall, we make the following \textbf{contributions}:
\begin{itemize} 
    \item  We present a novel method to understand layers of neural networks. It uses a decoder to translate incomprehensible outputs of an entire layer or a subset thereof into a humanly understandable representation. It allows to trade comprehensibility and fidelity. It eases the interpretation of explanations that also allow deriving general statements upon model behavior.
\item We introduce a method dealing with artifacts created by auxiliary models (or proxies) through comparisons with adequate references. This includes evaluation of methods using novel sanity checks.
\item This journal article extends the conference version by enhancing all manuscript sections with new material. It newly adds a user study, showing that our method enables even non-experts to identify a rich set of concepts shown on images that are maintained (or not maintained) in the classification process and a more detailed presentation of results and related work. It contains new use-cases, such as assessing subsets of layer activations down to individual neurons. It also adds an extensive evaluation using occlusion of inputs introducing novel measures to compare concept-based methods using reconstructions with saliciency maps from GradCAM. The journal version also includes an extensive reflection (Discussion and Future Work section), an assessment of the impact of classifier performance on reconstructions, and a comparison against prototype-based methods.
\end{itemize}

\section{Related Work}
\begin{figure*}
  \centering
  \includegraphics[width=\linewidth]{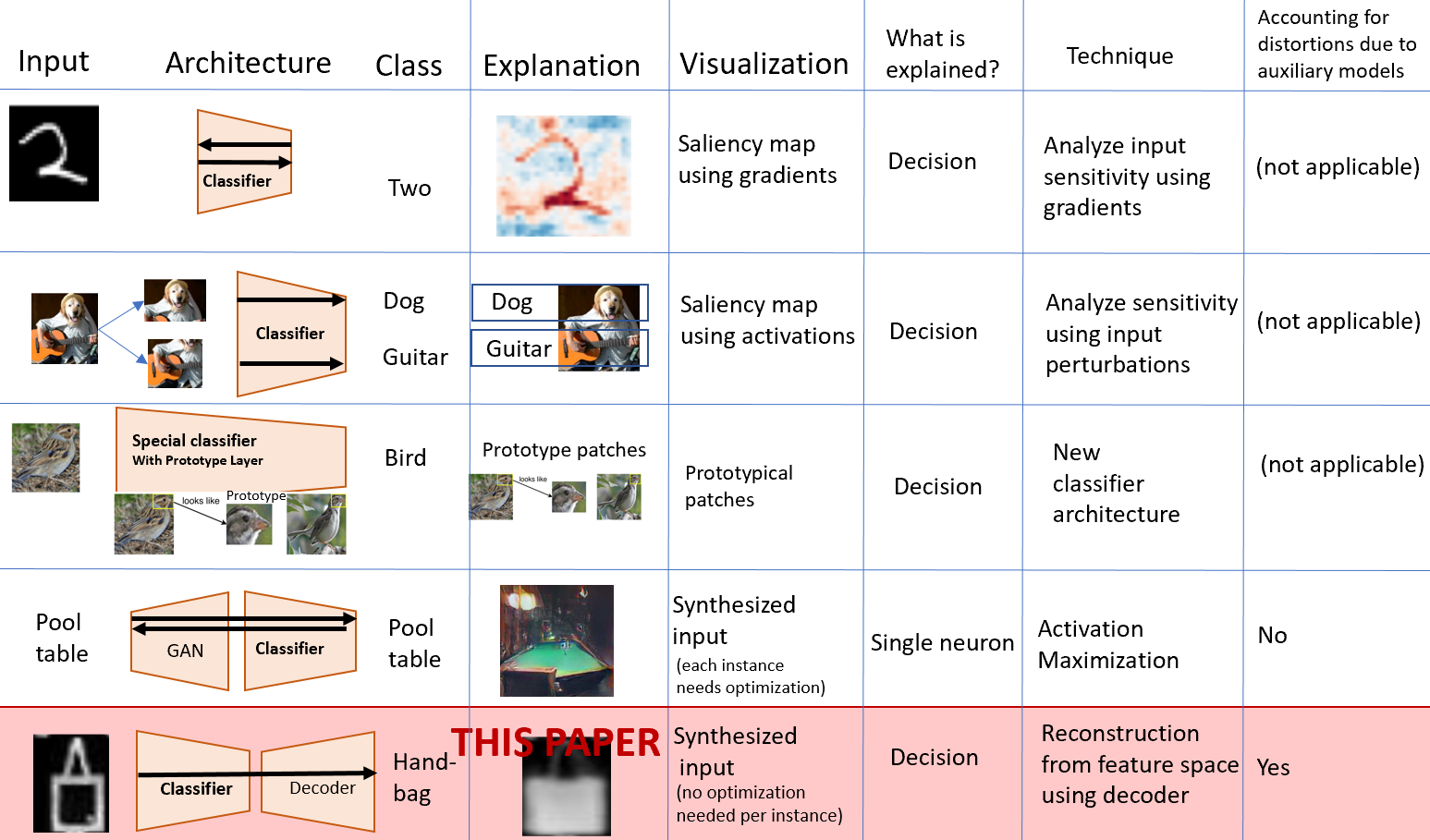}
  \caption{Method Overview. Figures are from cited papers.} \label{fig:met}
\end{figure*}

We first discuss approaches that allow us to visualize single features and understand model decisions summarized in Figure \ref{fig:met}.

We categorize explainability methods~ \cite{schneider2019pers} into methods that synthesize inputs (like ours and \cite{ngu16,yos15,gui19bl,aga20}) and methods that rely on saliency maps \cite{sim13} based on perturbation~ \cite{ribeiro2016should,zeil14} or gradients~ \cite{selvaraju2017grad,bach2015pixel}. Saliency maps show the feature importance of inputs, whereas synthesized inputs often show higher level representations encoded in the network. Perturbation-based methods include occlusion of parts of the inputs \cite{zeil14,wu2020exp} and investigating the impact on output probabilities of specific classes. Linear proxy models such as LIME~ \cite{ribeiro2016should} perform local approximations of a black-box model using simple linear models by also assessing modified inputs. Saliency maps \cite{sim13} highlight parts of the inputs that contributed to the decision. Many explainability methods have been under scrutiny for failing sanity checks \cite{adebayo2018sanity} and being sensitive to factors not contributing to model predictions \cite{kin19} or adversarial perturbations \cite{gho19}. Even if many of them might nevertheless be considered helpful and there are attempts to remedy these issues \cite{yeh19}, explanations are still fairly trivial. That is, mere highlighting does not provide any insights into how (input) information is processed and how it is encoded in the network \cite{rud19}. For those methods that show gradients (or a function of the gradients), one primarily sees how (very) small changes would impact the output, e.g., red might improve confidence in a class and blue reduce. Still, it is generally unclear whether large changes would still yield the behavior as suggested in the explanation, i.e., either an increase or a decrease. This is because gradients are only valid locally and might give no information on function behavior far from the point they are computed.


We anticipate that our work is less sensitive to targeted, hard to notice perturbations \cite{gho19} as well as translations or factors not impacting decisions \cite{kin19} since we rely on encodings of the classifier. Thus, explanations only change if these encodings change, i.e., if the classifier is sensitive to the perturbations. The idea to evaluate explanations on downstream tasks is not new, however a comparison to a ``close'' baseline like our \emph{RefAE} is. Our ``evaluation classifier'' using only explanations (without inputs) is more suitable than methods like \cite{sch20ref} that use explanations together with inputs in a more complex, non-standard classification process. Using inputs and explanations for the evaluation classifier is diminishing differences in evaluation outcomes since a network might extract missing information in the explanation from the input. So far, inputs have only been synthesized to understand individual neurons \cite{ngu16,bar20}, where the pioneering work \cite{ngu16} used activation maximization in an optimization procedure. The idea is to identify inputs that maximize the activation of a given neuron. This is similar to the idea to identify samples in the input that maximize neuron activation. \cite{ngu16} uses a (pre-trained) GAN on natural images relevant to the classification problem. It identifies through optimization the latent code that when fed into the GAN results in a more or less realistic looking image that maximally activates a neuron.  \cite{yos15} uses regularized optimization as well, yielding artistically more interesting but less recognizable images. Regularized optimization has also been employed in other forms of explanations of images, e.g., to make humans understand how they can alter visual inputs such as handwriting for better recognizability by a CNN \cite{sch20hu}. \cite{aga20} uses a GAN to replace removed input features, e.g., through cropping of parts of the input image, using realistic in-painting in the context of explainability.
\cite{gui19bl} trains an adversarial auto-encoder (AAE) on the training data. The idea is to generate similar samples to an input $X$ using the AAE by distorting the latent encoding of $X$. The generated samples are labeled using the original classifier. Then, an approximate model using a decision tree is trained using the labeled data. This also allows to obtain contrastive explanations. Rather than using an AAE one might use \emph{ClaDec} to generate similar samples. This might be even more appropriate since these reconstructions would be based on the latent space of the classifier instead of the latent space of the AAE. That is, \emph{ClaDec} focuses more strongly on differences in generated samples that are also relevant to classification. In contrast to our work, \cite{gui19bl} does not propose to compare to a reference, i.e., the generated samples might exhibit distortions stemming from the AAE that are misleading. \cite{van20} uses a variational AE for contrastive explanations. They use distances in latent space to identify samples that are closest to a sample $X$ of class $Y$ but actually classified as $Y'$. \emph{ClaDec} might also be used to this end to work directly using the latent space of the classifier rather than the one of the encoder from a separate AE.

\cite{kim17,ghor19} investigate high level concepts that are relevant to a specific decision. DeepLift \cite{shr17} compares activations to a reference and propagates them backward. Defining the reference is non-trivial and domain specific. \cite{koh2017} estimates the impact of individual training samples. \cite{liu19tow} discusses how to explain variational AEs using gradient-based methods. \emph{ClaDec} could also be used to explain AEs. \cite{che18,li18de} propose new network architectures that are based on encoding prototypes. In contrast to other methods that allow for post-hoc explanations, explainability is built into the model. The reasoning process is based on using dedicated layers/convolutions that encode prototypical patches/samples. While \cite{che18} achieves good classification performance, it imposes constraints on the classifier design that can lead to inferior  classifier performance. In contrast, our method is not imposing any constraints and is universally applicable. We also do not explicitly learn parts or patches like \cite{che18}. \cite{wu2020exp} aims at global explanations. They use a two stage process using occlusion of inputs and ad-hoc semantic analysis. The visualization of class concepts (Figure 3 in \cite{wu2020exp}) allows to draw some conclusions on network behavior, but appears highly distorted since it is purely based on activation maximization. \cite{raf20} used pre-defined concepts such as color and class association to classify neurons. The idea to focus on individual neurons has been criticized \cite{fong18} since concepts are often not encoded by a single neuron but by groups. Our work aims more at the question ``What concepts are relevant given layer activations?''. 
Other works investigated the usage of domain knowledge rather than high-level concepts \cite{conf19}.

Denoising AEs are well established \cite{vin10,du16}. They can be used to remove noise from images, reconstruct images and leverage data in an unsupervised manner \cite{vin10,du16}. Ideas to combine unsupervised learning approaches to remove noise and supervised learning by extending loss functions have been presented since the early 90's \cite{dec93}. In the context of explanations, AEs are also common \cite{van20,gui19bl,qi19co}, e.g., \cite{qi19co} used an AE with skip-connections for saliency map predictions. However, the encoder is fixed in our work, i.e., only we use the classifier directly as encoder. Furthermore, only in our work an AE is trained to identify distortion within the explanation process.

\section{Method and Architecture}
In our local, model-agnostic post-hoc explanation method, an explanation consists of three images $X,\hat{X}_E$ and $\hat{X}_R$, which are compared among each other as illustrated in the right of Figure \ref{fig:arch}. 
The \emph{ClaDec} architecture is shown on the top portion of Figure \ref{fig:arch}. The entire classifier has been trained beforehand to optimize classification loss. Its parameters remain unchanged during the explanation process. To explain layer $L$ of the classifier (= encoder) for an input $X$, we use the activations of the entire layer $L(X)$ (or a subset $S(X) \subseteq L(X)$). The decoder is trained to optimize the reconstruction loss given the activations $L(X)$ with respect to the original inputs $X$. The \emph{RefAE} architecture is identical to \emph{ClaDec}. It differs only in the training process and the objective. For the reference AE, the encoder and decoder are trained jointly to optimize the reconstruction loss of inputs $X$. In contrast, the encoder is treated as fixed in \emph{ClaDec}. Once the training of all components is completed, explanations can be generated without further need for optimization. That is, for an input $X$, \emph{ClaDec} computes the reconstruction $\hat{X}_E$ serving together with the original input and the reconstruction from \emph{RefAE} as the explanation.

However, comparing the reconstruction $\hat{X}_E$ to the input $X$ may be difficult and even misleading since the decoder can introduce distortions. Therefore, it is unclear, whether the differences between the input and the reconstruction originate from the encoding of the classifier or the inherent limitations of the decoder. Thus, we propose to use both the \emph{RefAE} (capturing unavoidable limitations of the model or data) and \emph{ClaDec} (capturing model behavior). The evaluation proceeds by comparing the reconstructed ``reference'' from \emph{RefAE}, the explanation from \emph{ClaDec} and the input. Only differences between the input and the reconstruction of \emph{ClaDec} that do not occur in the reconstruction \emph{RefAE} can be attributed to the classifier. In case, reconstructions by the \emph{RefAE} are (almost) identical to the original images, it suffices to compare only the reconstruction $\hat{X}_E$ by \emph{ClaDec} to the input $X$.

\begin{figure}
  \centering
  \includegraphics[width=0.85\linewidth]{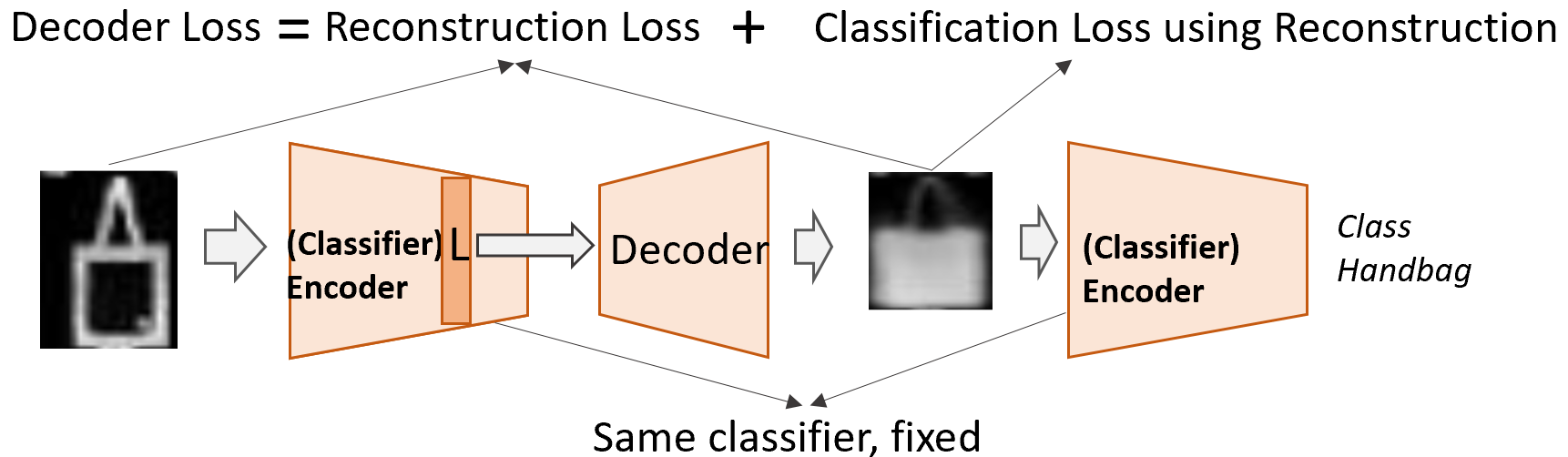}
  \caption{Extension of the \emph{ClaDec} architecture. The decoder is optimized for reconstruction and classification loss} \label{fig:deco}
\end{figure}

The resulting reconstructions might be easy to interpret, but in some cases it might be preferable to allow for explanations that are more fidel, i.e., capturing more aspects of the model that should be explained. Figure \ref{fig:deco} shows an extension of the base architecture of \emph{ClaDec} (Figure \ref{fig:arch}) using a second loss term for the decoder training. It is motivated by the fact that \emph{ClaDec} seems to yield reconstructions that capture more aspects of the input domain than of the classifier. 

More formally, for an input $X$, a classifier $C$ (to be explained) and a layer $L$ to explain, let $L(X)$ be the activations of layer $L$ for input $X$, and $Loss(C(X),Y)$ the classification loss of $X$ depending on the true classes $Y$. Let $S(X) \subseteq L(X)$ be the subset of neurons to explain, i.e., $S(X)=L(X)$ implies neurons of the entire layer should be explained. The decoder $D$ transforms the representation $S(X)$ into the reconstruction $\hat{X}$.
For \emph{ClaDec} the decoder loss is:
\begin{equation} 
\begin{aligned}	
Loss_{ClaDec}(X)&:= (1-\alpha)\cdot \sum_i (X_i-\hat{X}_{E,i})^2 + \alpha\cdot Loss(C(\hat{X}_{E}),Y)\\
&\text{ with } \hat{X}_{E}:=D(S(X)) \text{ and }\alpha \in[0,1] 
 \end{aligned}	\label{eq:alpha}
\end{equation}
The trade-off parameter $\alpha$ allows controlling whether reconstructions $\hat{X}_{E}$ are more similar to inputs with which the domain expert is more familiar, or  reconstructions that are more shaped by the classifier and, thus, they might look more different than training data a domain expert is familiar with. For reconstructions $X_{R}$ of \emph{RefAE} the loss is only the reconstruction loss: 
$$Loss_{RefAE}(X):= \sum_i (X_i-\hat{X}_{R,i})^2 $$
\section{Theoretical Motivation of \emph{ClaDec}} 
\begin{figure}
  \centering
  \includegraphics[width=0.8\linewidth]{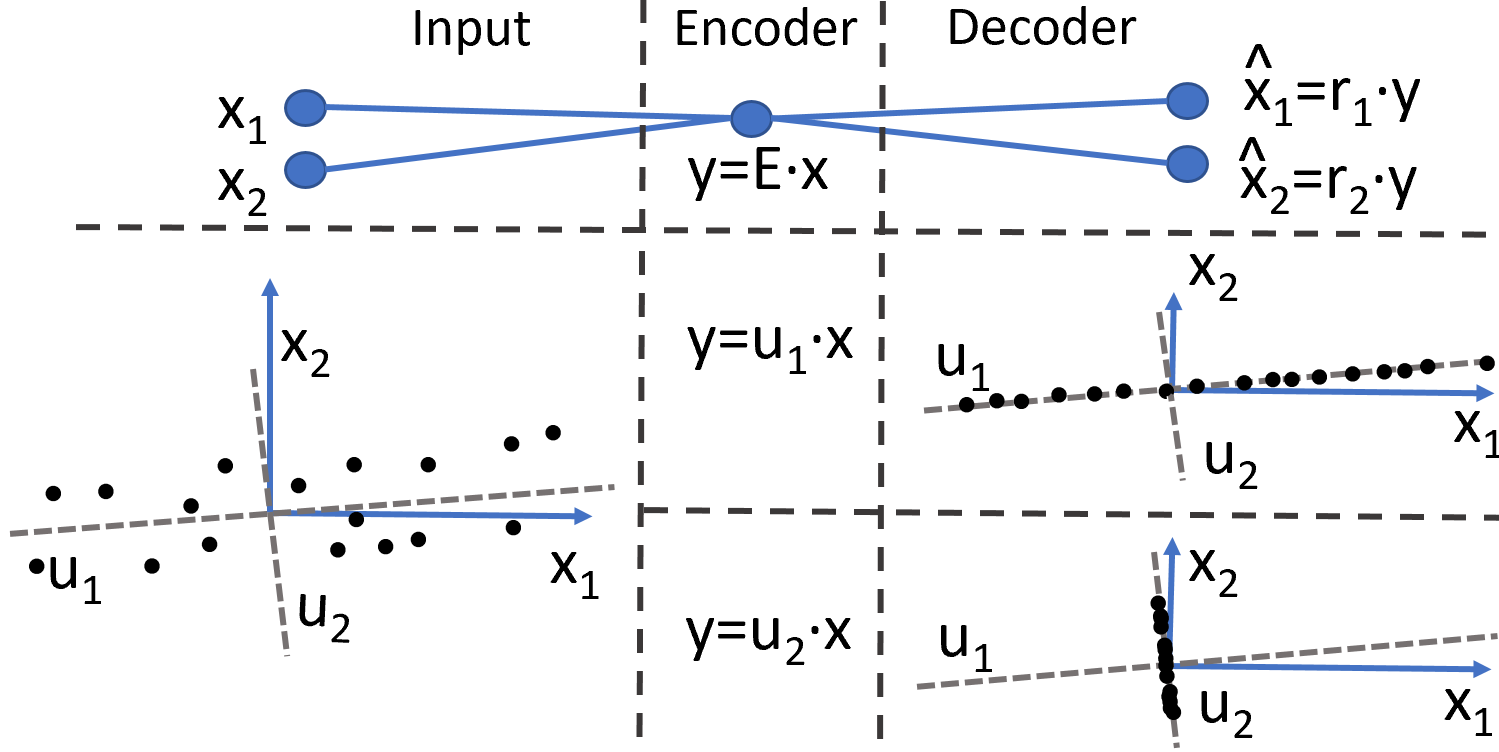}
  \caption{An AE with optimal encoder $y=u_1\cdot x$ (and decoder) captures more information than any other encoder. But a regression/classification model serving as encoder, e.g., $y=u_2\cdot x$, combined with an optimized decoder, might capture some input attributes more accurately, e.g., $x_2$.} \label{fig:the}
\end{figure}

We provide rationale for reconstructing explanations using a decoder from a layer of a classifier that should be explained, and comparing it to the output of a conventional AE, i.e., \emph{RefAE} (see Figure \ref{fig:arch}). AEs perform a transformation of inputs to a latent space and then back to the original space. This comes with information loss on the original inputs because reconstructions are typically not identical to inputs.\footnote{It may appear that this information loss is due to forcing high-dimensional data to be represented in a low dimensional space. However, as claimed in \cite{goo16}(p.505), a non-linear encoder and decoder (theoretically) only require a single dimension to encode arbitrary information without any loss. The deeper mathematical reason is that a dimension $d$ is a real number, i.e., $d \in \mathbb{R}$ and real numbers are uncountable infinite. Thus, there are (more than) enough options to encode an infinite amount of inputs. Therefore, information loss is more a failure of the model to encode and decode the input error free using just one dimension.} 
To provide intuition, we focus on a simple architecture with a linear encoder (consisting of a linear model that should be explained), a single hidden unit and a linear decoder as depicted in Figure \ref{fig:the}.
An AE, i.e., the reference AE  \emph{RefAE}, aims to find an encoding vector $E$ and a reconstruction vector $R$, so that the reconstruction $\hat{x}=R\cdot y$ of the encoding $y=E\cdot x$ is minimal using the L2-loss:
$$\min_{R,E} ||x-R\cdot E\cdot x||^2$$
The optimal solution which minimizes the reconstruction loss stems from projecting onto the eigenvector space (as given by a Principal Component Analysis) \cite{bal89}. Given that there is just a single latent variable, the optimal solution for $W=R\cdot E$ is the first eigenvector $u_1$. This is illustrated in Figure \ref{fig:the} in the upper part with $y=u_1\cdot x$. For \emph{ClaDec} the goal is to explain a linear regression model $y=E\cdot x$. The vector $E$ is found by solving a regression problem. We fit the decoder $R$ to minimize the reconstruction loss on the original inputs given the encoding:
$$\min_{R} ||x-R\cdot y||^2 \text{ with } y=E\cdot x$$ The more similar the regression problem is to the encoding problem of an AE, the more similar are the reconstructions. Put differently, the closer $E$ is to $u_1$, the lower the reconstruction loss and the more similar are the optimal reconstructions for the reference AE and \emph{ClaDec}. Assume that $E$ differs strongly from $u_1$, e.g., the optimal solution to the regression problem is the second eigenvector $y=u_2\cdot x$. This is shown in the lower part of Figure \ref{fig:the}. When comparing the optimal reconstruction of the \emph{RefAE}, i.e., using $y=u_1x$, and the illustrated reconstruction of \emph{ClaDec}, i.e., using $y=u_2x$, it becomes apparent that for the optimal encoding $y=u_1x$ the reconstructions of both coordinates $x_1$ and $x_2$ are fairly accurate on average. In contrast, using $y=u_2x$, coordinate $x_2$ is reconstructed more accurately (on average), whereas the reconstruction of $x_1$ is mostly poor.

Generally, this suggests that a representation obtained from a model (trained for some task) captures less information of the input than an encoder optimized for reconstructing inputs. But aspects of inputs relevant to the task should be captured relatively in more detail than those irrelevant. Reconstructions from \emph{ClaDec} should show more similarity to original inputs for attributes relevant to classification and less similarity for irrelevant attributes. But, overall reconstructions from the classifier will show less similarity to inputs than those of an AE. Our fidelity assessment builds on this idea by proclaiming that reconstructions from an AE, i.e., \emph{RefAE}, capture more information on inputs (measured using reconstruction loss) than those from \emph{ClaDec}, but are less suitable for classification (measured using accuracy on a classifier trained on explanations).

\section{Assessing Comprehensibility and Fidelity} \label{sec:san} 

\textbf{Fidelity} is the degree to which an explanation captures model behavior. That is, a ``fidel'' explanation captures the decision process of the model accurately.

The proposed evaluation (also serving as a sanity check) uses the rationale that fidel explanations for decisions of a well-performing model should help performing the task the model addresses. That is, learning from explanations of a well-performing model should also lead to a well-performing model for the same task. Concretely, training a new classifier $C^{ClaDec}_{eval}$ on explanations should yield a better performing classifier than relying on reconstructed inputs only. 
That is, we train a baseline classifier $C^{RefAE}_{eval}$ on the reconstructions $\hat{X}_R$ of the \emph{RefAE} and a second classifier with identical architecture $C^{ClaDec}_{eval}$ on reconstructions $\hat{X}_E$ from \emph{ClaDec}. The latter classifier should achieve higher accuracy. This is a much stronger requirement than the common sanity check demanding that explanations must be valuable to perform a task better than a ``guessing'' baseline. More formally, we compare the accuracy $Acc(C^{ClaDec}_{eval})$ of the model $C^{ClaDec}_{eval}$ trained on reconsructions $\hat{X}_E$ and that of a model $C^{RefAE}_{eval}$ trained on reconstructions $\hat{X}_R$ of the \emph{RefAE}. Thus, as a (proxy) measure for fidelity $\Delta$Acc, we use $$\Delta Acc:=Acc(C^{ClaDec}_{eval})-Acc(C^{RefAE}_{eval})$$

One must be careful that explanations do not contain additional external knowledge (not present in the inputs or training data) that helps solving the task. For most methods, including ours, this holds. It is not obvious that training on explanations improves classification performance compared to training on inputs that are more accurate reconstructions of the original inputs. Improvements seem only possible if an explanation is a more adequate representation to solve the problem. Formally, we measure the similarity between the reconstructions $\hat{X}_R$ (using \emph{RefAE}) and $\hat{X}_E$ (of \emph{ClaDec}) with the original inputs $X$. We show that explanations (from \emph{ClaDec}) bear less similarity with original inputs than reconstructions from \emph{RefAE}. Still, training on explanations $\hat{X}_E$ only yields classifiers with better performance than on the more informative outputs $\hat{X}_R$ from \emph{RefAE}.

\noindent
\textbf{Comprehensibility} is the degree to which the explanation is human-understandable. In our case, this means whether a person can make sense of the concepts depicted in the explanation.  We build upon the intuitive assumption that a human can better and more easily comprehend explanations made of concepts that she is more familiar with. We argue that a user is more familiar with real-world phenomena and concepts as captured in the training data than possibly unknown concepts captured in representations of a neural network. This implies that more similar explanations to the training data are more comprehensible than those with strong deviations from the training data. Therefore, we quantify comprehensibility of a reconstruction $\hat{X}_E$ by measuring the distance to the original input $X$, i.e., the reconstruction loss $||X-\hat{X}_E||$. If reconstructions show fidelitous but non-intuitive concepts (high reconstruction loss) then a user can experience difficulties in making sense of the explanation. In contrast, a trivial explanation (showing the unmodified input) is easy to understand but it will not reveal any insights into the model behavior, i.e., it lacks fidelity. We consider the best ``trivial'' explanation that can be obtained through a reconstruction process that of the \emph{RefAE}. We discuss the reconstruction loss of the reconstructions $\hat{X}_E$ from \emph{ClaDec} with respect to that of the \emph{RefAE}. That is, for a test dataset $D=\{(X,Y)\}$, we report separately  the reconstruction loss for \emph{RefAE} and \emph{ClaDec} and the difference of these losses: $$\Delta Rec:=\frac{1}{|D|}\sum_{X,Y \in D}||X-\hat{X}_E||-||X-X_R||$$


\section{Evaluation}
We perform a qualitative and quantitative evaluation including a user study focusing on image classification using CNNs. We perform the following experiments: 
\begin{itemize}
    \item Explaining multiple layers for correct and incorrect classifications (Section \ref{sec:expCoIn})     
    \item Varying the fidelity and comprehensibility trade-off (Section \ref{sec:fidel})    
    \item Impact of encoder training and accuracy on explanations (Section \ref{sec:expAE})
    \item Using subsets of layer activations down to individual neurons (Section \ref{sec:layAct})
    \item Comparing \emph{ClaDec} explanations to class prototypes learnt in \cite{li18de} (Section \ref{sec:proto})
    \item Investigating impact of occluding parts of the input (Section \ref{sec:occ})
    \item A user study with non-experts asking them to make sense of explanations (Section \ref{sec:huExp})
\end{itemize}

\begin{figure}
  \centering
  \includegraphics[width=\linewidth]{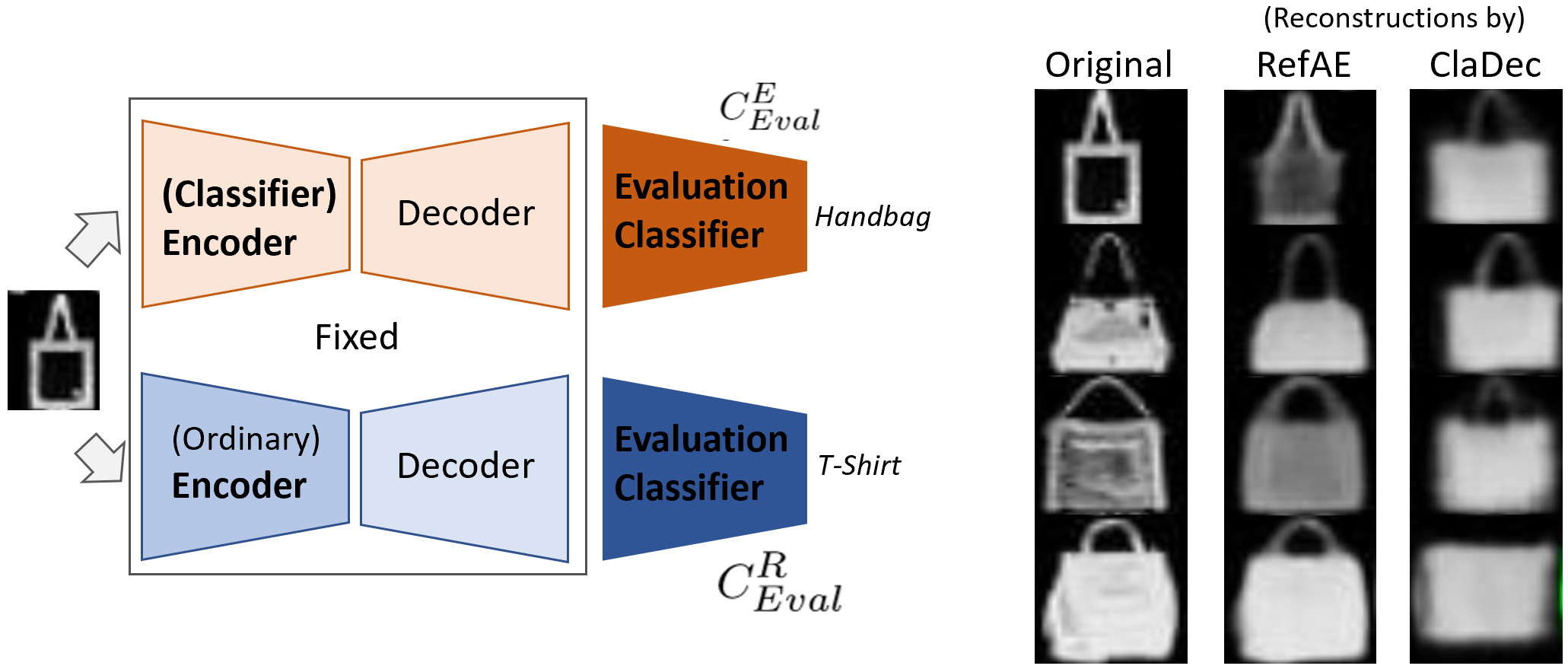} 
  \caption{Left panel: Evaluation setup using a dedicated evaluation classifier for \emph{ClaDec} and \emph{RefAE}. Right panel: Comparison of original inputs and reconstructions using the FC layer of the encoder for handbags. \emph{RefAE} and \emph{ClaDec} both do not reconstruct detailed textures. The classifier does not rely on gray tones, which are captured by \emph{RefAE}. It uses prototypical shapes.} \label{fig:recFC} %
\end{figure}

\noindent\textbf{Setup}: The decoder follows a standard design, i.e., using 5x5 deconvolutional layers. For the classifier and encoder we used the same architecture, i.e., a VGG-11 and ResNet-10. Architectures are shown in Figure \ref{tab:arch}. For ResNet-10 we reconstructed after each block. For VGG-11 after a ReLU unit associated with a conv layer. The same classifier architecture (but trained with different input data) serves as encoder in \emph{RefAE}, classifier in \emph{ClaDec} and for classifiers used for evaluation of reconstructions, i.e., classifier $C^{ClaDec}_{Eval}$ (for assessing \emph{ClaDec}) and  $C^{RefAE}_{Eval}$ (for \emph{RefAE}) shown in the left panel of Figure \ref{fig:recFC}. We report the validation accuracy ``Acc $C^{ClaDec}_{Eval}$'' and ``Acc $C^{RefAE}_{Eval}$'' of these classifiers.
Code is available at \url{https://github.com/JohnTailor/ClaDec}.\\
 \begin{table}
 	\vspace{-6pt}
 	\begin{center}
 		\scriptsize
 		\setlength\tabcolsep{2.5pt}
	\centering
 		\begin{tabular}{| l | l| l|l| }\hline
 			\multicolumn{2}{|c|}{VGG-style Encoder} & \multicolumn{2}{c|}{Decoder}\\  \hline
			Type/Stride & Filter Shape &Type/Stride& Filter Shape \\  \hline
 			C/s2     & $3\tiny{\times} 3 \tiny{\times} 1 \tiny{\times} 32$ & FC     & nClasses  \\ \hline
 			C/s2     & $3\tiny{\times} 3 \tiny{\times} 32 \tiny{\times} 64$ & DC/s2     & $3\tiny{\times} 5 \tiny{\times} 5 \tiny{\times} 256$  \\ \hline
 			C/s1     & $3\tiny{\times} 3 \tiny{\times} 64 \tiny{\times} 128$ & DC/s2     & $3\tiny{\times} 5 \tiny{\times} 5 \tiny{\times} 128$  \\ \hline
 			C/s2     & $3\tiny{\times} 3 \tiny{\times} 128 \tiny{\times} 128$ & DC/s2     & $3\tiny{\times} 5 \tiny{\times} 5 \tiny{\times} 64$  \\ \hline
 			C/s1     & $3\tiny{\times} 3 \tiny{\times} 128 \tiny{\times} 256$ & DC/s2     & $3\tiny{\times} 5 \tiny{\times} 5 \tiny{\times} 32$  \\ \hline
 			C/s2 & $3\tiny{\times} 3 \tiny{\times} 256 \tiny{\times} 256$ &     DC/s2 & $3\tiny{\times} 5 \tiny{\times} 5 \tiny{\times} 1$  \\ \hline
            C/s1 & $3\tiny{\times} 3 \tiny{\times} 256 \tiny{\times} 512$ &     &  \\ \hline
            C/s2 & $3\tiny{\times} 3 \tiny{\times} 512 \tiny{\times} 512$ &     &  \\ \hline
 			FC & $256 \tiny{\times} $nClasses &  &    \\ \hline
 			Softmax/s1 & Classifier &   &  \\ \hline
 			\end{tabular}
` 	
 	\end{center}
 	\caption{Encoder/Decoder, where ``C'' is a convolution, ``DC'' a deconv; a BatchNorm and a ReLu layer follow each ``C'' layer; a ReLu layer follows each ``DC'' layer}  \label{tab:arch} 
 	\vspace{-6pt}
 \end{table}

Note that the decoder architecture varies depending on which layer is to be explained. The original architecture allows to either obtain reconstructions from the last convolutional layer or the fully connected layer. For a lower layer, the highest deconvolutional layers from the decoder have to be removed, so that the reconstructed image $\hat{X}$ has the same width and height as the original input $X$. We employed three datasets namely Fashion-MNIST, MNIST and CIFAR-100. Fashion-MNIST consists of 70000 28x28 images of clothing stemming from 10 classes that we scaled to 32x32. MNIST of 60000 digits and CIFAR-100 of 60000 objects in color. 10000 samples are used for testing. We train all models for reconstruction using the Adam optimizer for 64 epochs, i.e., \emph{RefAE} and the decoder of \emph{ClaDec}. The classifier serving as encoder in \emph{ClaDec} as well as the classifiers used for evaluation for SGD were trained using SGD for 64 epochs starting from a learning rate of 0.1 that was decayed twice by 0.1. We conducted 5 runs for each reported number. We show both averages and standard deviations. The classifier performance for each of the dataset and architecture is comparable to those in other papers without data augmentation, i.e., for MNIST we achieved mean accuracy above 99\%, for FashionMNIST above 92\%, and for CIFAR-100 above 45\% for both architectures.

\begin{figure}
  \centering
  \includegraphics[width=1.0\linewidth]{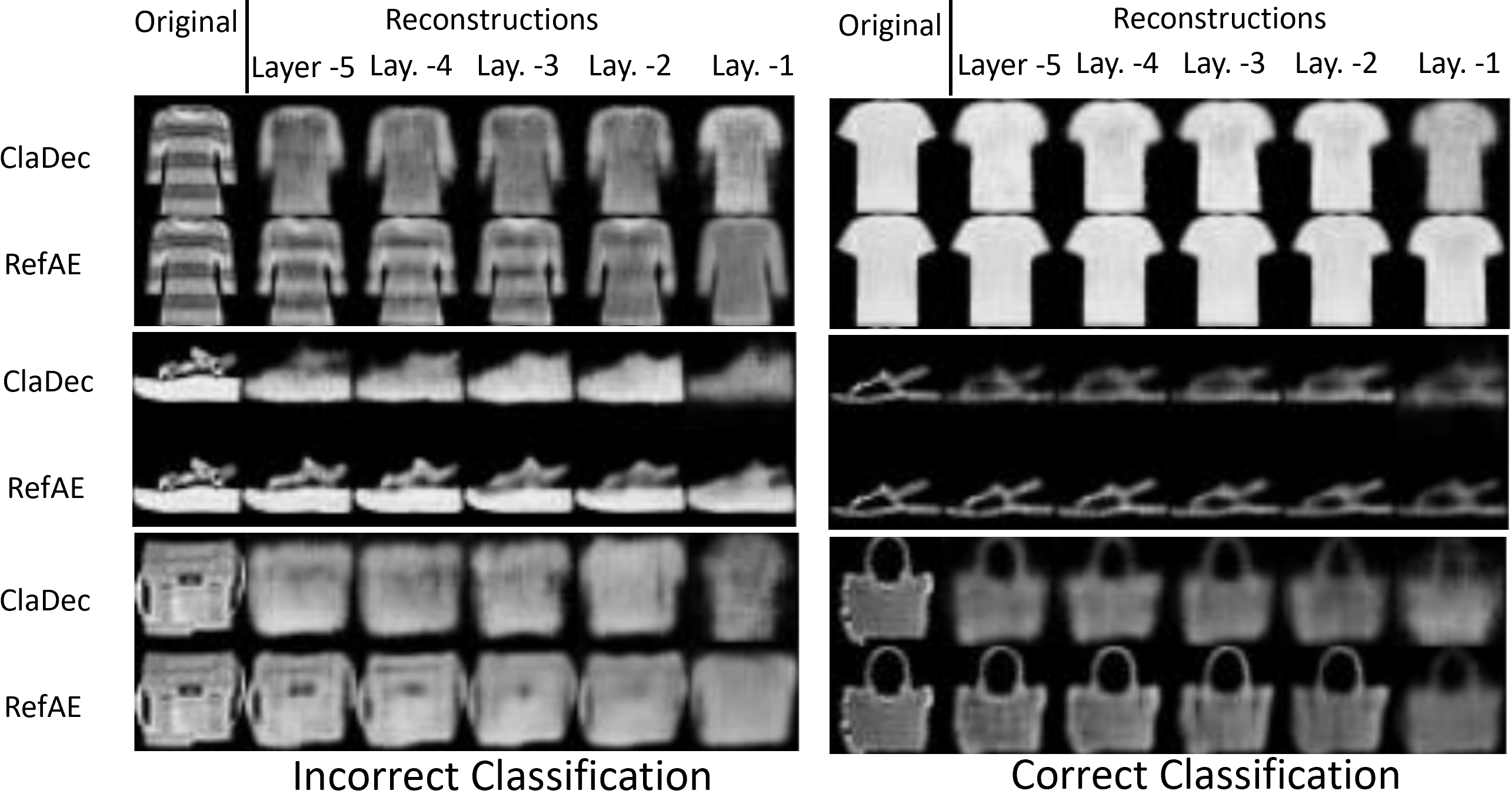} 
  \caption{Comparison of original inputs and reconstructions using multiple layers of the encoder. For incorrect samples it shows a gradual transformation into another class. Differences between \emph{RefAE} and \emph{ClaDec} increase with each layer \label{fig:cowr}} 
\end{figure}

\subsection{Explaining layers for correct and incorrect classifications} \label{sec:expCoIn}


\begin{figure}
  \centering
  \includegraphics[width=\linewidth]{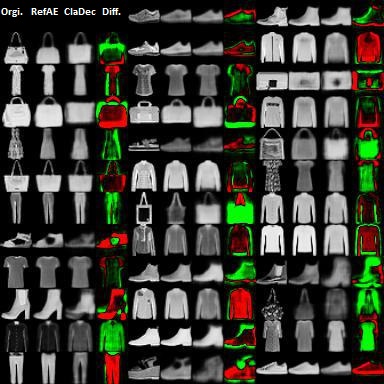}
  \caption{Comparison of original inputs and reconstructions using the last layer, i.e., FC, of the VGG encoder in Table  \ref{tab:arch}. Differences between reconstructions are shown in the last column. Red areas show where reconstructions from \emph{RefAE} are brighter than those from \emph{ClaDec}. Green shows the opposite. } \label{fig:show1} 
\end{figure}

Reconstructions based on \emph{RefAE} and \emph{ClaDec} for FashionMNIST are shown in Figures \ref{fig:recFC}, \ref{fig:cowr} and \ref{fig:show1}. Overall reconstructions from \emph{ClaDec} resemble more prototypical, abstract features and they allow to identify relevance of input details due to imprecise reconstruction (blurriness, change of shape) or complete absence of concepts such as textures or gray tones. That is, omitted concepts are not relevant, while for poor reconstructions, the (blurry) input parts are not relevant at a high level of detail to discriminate between classes. In Figure \ref{fig:recFC} the difference between reconstructions of \emph{RefAE} and \emph{ClaDec} are not the same as in a salience map, i.e., they do not indicate that a red part is relevant according to \emph{ClaDec} and a green part is not or vice versa. It only shows differences in pixel values. This is not directly translatable to relevance. The classifier (and thus the reconstruction from \emph{ClaDec}) can turn a fairly dark area in the input into a gray area or a fairly white area into a gray area. This primarily indicates that the color or gray tune is not important, but it does not mean that in one case importance of the pixel is low and in the other it is  high.  \\

\begin{figure}
   \centering
   \includegraphics[width=\linewidth]{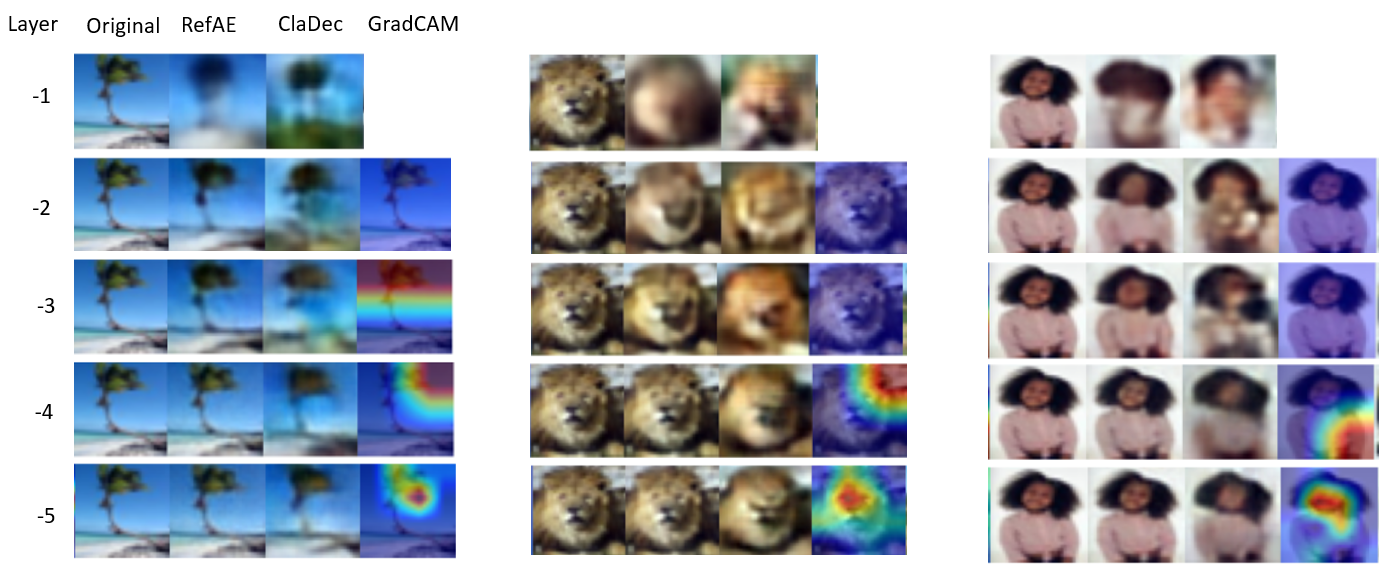} 
   \caption{Comparison of original and reconstructions and GradCAM for different layers using ResNet on CIFAR-100} \label{fig:resCi100} 
 \end{figure}

\begin{figure}
   \centering
   \includegraphics[width=\linewidth]{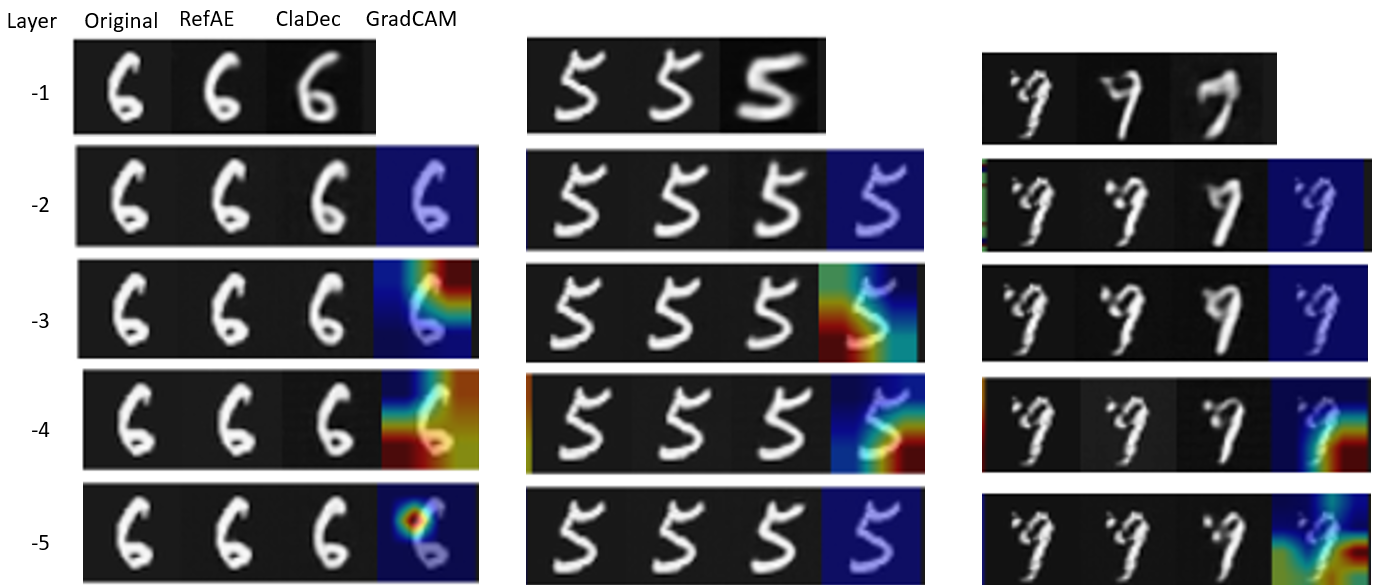} 
   \caption{Comparison of original and reconstructions and GradCAM for different layers using ResNet on MNIST} \label{fig:resMNIST} 
 \end{figure}
 
MNIST and CIFAR-100 exhibited similar behavior as Fashion-MNIST as can be seen in Figures \ref{fig:resCi100} and \ref{fig:resMNIST}. For CIFAR-100 analyzing explanations is more cognitively demanding since there are more classes, classes exhibit more diversity and reconstructions of both \emph{RefAE} and \emph{ClaDec} are of worse quality compared to the other two datasets.  
First of all, it should be noted that GradCAM does not give favorable results for layers close to the output since these layers lack spatial extent. For example, for the third last layer the spatial dimensions of activation layers is 2x2 requiring an upsampling by a factor of 16 for each dimension to the final output, which yields essentially a uniform attribution map of GradCAM. For the very last layer GradCAM yields meaningful results, i.e., for CIFAR-100 in Figure \ref{fig:resCi100} they show that the classifier focuses on the treetop as well as the girl's and lion's face. In fact, the situation is more delicate since GradCAM explanations do not clarify how the classifier perceives the highlighted pixels, i.e., does the classifier rely either on skin color or facial features such as nose or eyes or both?

For \emph{RefAE} reconstructions tend to lack details for CIFAR-100 (Figure \ref{fig:resCi100}), in particular for the last layer, where the number of dimensions is just 100 compared to 512 of the prior layer. However, the original input and \emph{RefAE} are overall still fairly similar. Comparing \emph{RefAE} and \emph{ClaDec} shows that both exhibit similar reconstructions for the two layers closest to the input with \emph{ClaDec} appearing slightly more blurry. For upper layers, reconstructions show partially semantic differences. For the tree, the shape is changed to be more prototypical for a palm tree. While the blue sky remains well visible, the ground does not appear as a white sandy beach including parts of the sea but it is altered from a more yellowish (sandy) ground to greenery. Compared to GradCAM, \emph{ClaDec} provides a more diversified understanding of the layers since the reconstructions give more information for each layer. There is agreement between the two methods that the treetop is highly relevant. \emph{ClaDec} shows that the tree shape is somewhat different for the input than for a typical sample, i.e., the collection of branches face more upwards in the original but more downward on the reconstruction (layers -2 and -3), which seems to be more characteristic (average) representation. Interestingly, the sandy beach is not well-reconstructed at the very top layer, whereas blue sky remains. This hints that the sandy ground and the ocean in the original do not lead the classifier to move towards a setting at a beach with sand and ocean. For the girl on the right the behavior is also interesting since starting from layer -3 upwards the reconstructions from \emph{ClaDec} appear to contain significant distortion in the center ultimately leading to wrong classification. That is, the very top does not show a black girl but a (white) woman. The girl's hair is changed from layer -4 to -3 from its fluffy appearance to a more common appearance (found more often in the training data). The change of skin color occurs from layer -3 to -2. This shows that the classifier is not picking up less common looks such as fluffy hair as a criterion for identification of the person and it highlights how the hair is transformed. For the lion, colors appear more vivid, i.e., brighter with larger hue, for \emph{ClaDec} showing how the more pale input deviates from the typical representation of colors. Also on the topmost layer the face of the lion is shrunk and more background is added, highlighting that such a setting is more common.\\
For MNIST (Figure \ref{fig:resMNIST}) the reconstructions of \emph{RefAE} are highly accurate. There are clearly visible differences in the top layer only for digit ``9''. For \emph{ClaDec} reconstructions for digits ``5'' and ``6'' are highly accurate except for the last layer, where more typical samples are reconstructed and sharpness also decreases for ``5''. The digit ``9'' is most interesting - it is also classified wrongly as ``7'' as can be seen well from the top layer reconstruction with the change already emerging at lower layers. At layer -3 the reconstruction still seems to point to a ``9'', with some uncommon remains of the ``big dot'' and the loop not being fully closed.  At layer -2 the opening of the loop increases somewhat and now the distinction between ``7'' and ``9'' is less clear. GradCAM explanations are not so insightful. \\

\begin{figure}
   \centering
   \includegraphics[width=0.8\linewidth]{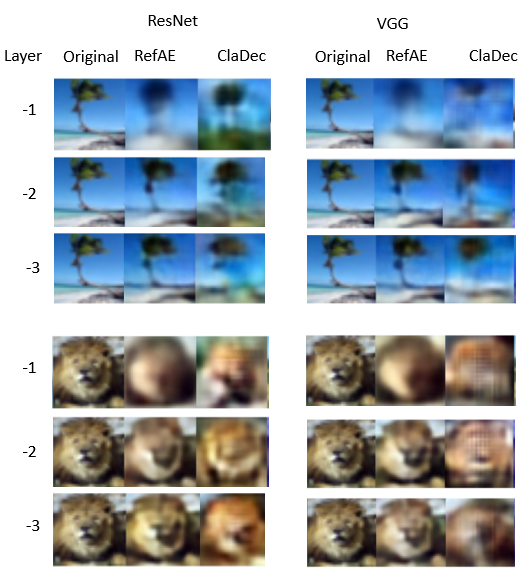} 
   \caption{Comparison of VGG and ResNet reconstructions on CIFAR-100} \label{fig:resVGG} 
 \end{figure}
 
We also compared reconstruction of ResNet and VGG shown in Figure \ref{fig:resVGG}. The reconstructions of the \emph{RefAE} are similar for both architectures, hinting that there is consistency -- at least for the reference. This is not obvious since the encoder architectures differ. However, for \emph{ClaDec} the reconstructions of the two classifiers show more variation. Quality of both, i.e., sharpness and shape characteristics, are similar with ResNet having arguably a slight advantage. But reconstructed concepts show differences. The ground for the beach differs strongly between the two as well as the shape of the treetop, while both maintain the blue sky. For the lion color tones differ also, with ResNet showing more saturated colors.

\begin{table}
\footnotesize
\begin{tabular}{| l|l|l|| l|l|l|l|}\hline
Layer&Rec Loss \emph{ClaDec}&Rec Loss \emph{RefAE}&$\Delta$Rec&Acc $C^{ClaDec}_{Eval}$&Acc $C^{RefAE}_{Eval}$&$\Delta$Acc  \\ \hline
\multicolumn{7}{|c|}{FashionMNIST} \\ \hline
-1&0.24\text{\tiny{$\pm$0.0183}}&0.038\text{\tiny{$\pm$0.0003}}&0.203&0.905\text{\tiny{$\pm$0.0003}}&0.917\text{\tiny{$\pm$0.0019}}&-0.011\\ \hline
-2&0.109\text{\tiny{$\pm$0.0047}}&0.017\text{\tiny{$\pm$0.0004}}&0.092&0.918\text{\tiny{$\pm$0.0021}}&0.921\text{\tiny{$\pm$0.0011}}&-0.003\\ \hline
-3&0.062\text{\tiny{$\pm$0.0015}}&0.013\text{\tiny{$\pm$0.0005}}&0.049&0.92\text{\tiny{$\pm$0.0022}}&0.919\text{\tiny{$\pm$0.0023}}&0.002\\ \hline
-4&0.042\text{\tiny{$\pm$0.0007}}&0.003\text{\tiny{$\pm$0.0004}}&0.039&0.925\text{\tiny{$\pm$0.0016}}&0.917\text{\tiny{$\pm$0.0029}}&0.008\\ \hline
-5&0.038\text{\tiny{$\pm$1e-04}}&0.002\text{\tiny{$\pm$0.0002}}&0.036&0.927\text{\tiny{$\pm$0.0013}}&0.915\text{\tiny{$\pm$0.0005}}&0.012\\ \hline
\multicolumn{7}{|c|}{MNIST} \\ \hline
-1&0.374\text{\tiny{$\pm$0.0042}}&0.028\text{\tiny{$\pm$0.0005}}&0.346&0.993\text{\tiny{$\pm$0.0004}}&0.989\text{\tiny{$\pm$0.0009}}&0.004\\ \hline
-2&0.139\text{\tiny{$\pm$0.0022}}&0.012\text{\tiny{$\pm$0.0005}}&0.127&0.995\text{\tiny{$\pm$0.0005}}&0.99\text{\tiny{$\pm$0.0003}}&0.005\\ \hline
-3&0.074\text{\tiny{$\pm$0.0013}}&0.01\text{\tiny{$\pm$0.0004}}&0.064&0.995\text{\tiny{$\pm$0.0002}}&0.992\text{\tiny{$\pm$0.0006}}&0.003\\ \hline
-4&0.038\text{\tiny{$\pm$0.0013}}&0.003\text{\tiny{$\pm$0.0001}}&0.035&0.995\text{\tiny{$\pm$0.0006}}&0.992\text{\tiny{$\pm$0.0006}}&0.003\\ \hline
-5&0.029\text{\tiny{$\pm$0.0004}}&0.002\text{\tiny{$\pm$0.0002}}&0.027&0.995\text{\tiny{$\pm$0.0003}}&0.994\text{\tiny{$\pm$0.0004}}&0.001\\ \hline
\multicolumn{7}{|c|}{CIFAR-100} \\ \hline
-1&0.16\text{\tiny{$\pm$0.0}}&0.12\text{\tiny{$\pm$0.001}}&0.04&0.31\text{\tiny{$\pm$0.003}}&0.37\text{\tiny{$\pm$0.002}}&-0.05\\ \hline
-2&0.11\text{\tiny{$\pm$0.001}}&0.06\text{\tiny{$\pm$0.002}}&0.05&0.41\text{\tiny{$\pm$0.004}}&0.37\text{\tiny{$\pm$0.007}}&0.05\\ \hline
-3&0.13\text{\tiny{$\pm$0.002}}&0.05\text{\tiny{$\pm$0.001}}&0.08&0.43\text{\tiny{$\pm$0.002}}&0.37\text{\tiny{$\pm$0.009}}&0.06\\ \hline
-4&0.1\text{\tiny{$\pm$0.001}}&0.02\text{\tiny{$\pm$0.002}}&0.09&0.48\text{\tiny{$\pm$0.002}}&0.41\text{\tiny{$\pm$0.005}}&0.07\\ \hline
-5&0.08\text{\tiny{$\pm$0.001}}&0.01\text{\tiny{$\pm$0.0}}&0.07&0.49\text{\tiny{$\pm$0.004}}&0.43\text{\tiny{$\pm$0.003}}&0.06\\ \hline
\end{tabular}
\caption{Explaining layers for VGG: \emph{ClaDec} has larger reconstruction loss but the evaluation classifier has higher accuracy on \emph{ClaDec}'s reconstructions } \label{tab:lay}
\end{table}

\begin{table*}
\footnotesize
\begin{tabular}{| l|l|l|| l|l|l|l|}\hline
Layer&Rec Loss \emph{ClaDec}&Rec Loss \emph{RefAE}&$\Delta$Rec&Acc $C^{ClaDec}_{Eval}$&Acc $C^{RefAE}_{Eval}$&$\Delta$Acc  \\ \hline
\multicolumn{7}{|c|}{FashionMNIST} \\ \hline
-1&0.1444\text{\tiny{$\pm$0.00338}}&0.0287\text{\tiny{$\pm$0.00048}}&0.1158&0.916\text{\tiny{$\pm$0.00248}}&0.922\text{\tiny{$\pm$0.00107}}&-0.006\\ \hline
-2&0.0222\text{\tiny{$\pm$0.00075}}&0.0006\text{\tiny{$\pm$0.00011}}&0.0216&0.9345\text{\tiny{$\pm$0.00115}}&0.9256\text{\tiny{$\pm$0.00212}}&0.0089\\ \hline
-3&0.0069\text{\tiny{$\pm$0.00056}}&0.0004\text{\tiny{$\pm$8e-05}}&0.0065&0.9353\text{\tiny{$\pm$0.00047}}&0.931\text{\tiny{$\pm$0.00035}}&0.0043\\ \hline
-4&0.0171\text{\tiny{$\pm$0.00487}}&0.0006\text{\tiny{$\pm$0.00012}}&0.0165&0.936\text{\tiny{$\pm$0.00122}}&0.9308\text{\tiny{$\pm$0.00111}}&0.0052\\ \hline
-5&0.0023\text{\tiny{$\pm$0.00069}}&0.0015\text{\tiny{$\pm$0.0005}}&0.0008&0.9362\text{\tiny{$\pm$0.0015}}&0.935\text{\tiny{$\pm$0.00096}}&0.0012\\ \hline

\multicolumn{7}{|c|}{MNIST} \\ \hline
-1&0.1722\text{\tiny{$\pm$0.00184}}&0.0194\text{\tiny{$\pm$0.00079}}&0.1528&0.9952\text{\tiny{$\pm$0.00098}}&0.9833\text{\tiny{$\pm$0.00226}}&0.0119\\ \hline
-2&0.0148\text{\tiny{$\pm$0.00025}}&0.0004\text{\tiny{$\pm$3e-05}}&0.0143&0.9964\text{\tiny{$\pm$0.00024}}&0.9955\text{\tiny{$\pm$0.00032}}&0.0009\\ \hline
-3&0.0045\text{\tiny{$\pm$0.00016}}&0.0002\text{\tiny{$\pm$0.00016}}&0.0043&0.9964\text{\tiny{$\pm$0.00019}}&0.9958\text{\tiny{$\pm$0.00022}}&0.0006\\ \hline
-4&0.0026\text{\tiny{$\pm$0.00039}}&0.0001\text{\tiny{$\pm$3e-05}}&0.0025&0.9963\text{\tiny{$\pm$0.00022}}&0.9962\text{\tiny{$\pm$0.00039}}&0.0001\\ \hline
-5&0.001\text{\tiny{$\pm$0.00015}}&0.0006\text{\tiny{$\pm$8e-05}}&0.0004&0.996\text{\tiny{$\pm$0.00036}}&0.9962\text{\tiny{$\pm$0.00044}}&-0.0002\\ \hline

\multicolumn{7}{|c|}{CIFAR-100} \\ \hline
-1&0.185\text{\tiny{$\pm$0.002}}&0.071\text{\tiny{$\pm$0.0017}}&0.114&0.389\text{\tiny{$\pm$0.0038}}&0.508\text{\tiny{$\pm$0.0027}}&-0.12\\ \hline
-2&0.057\text{\tiny{$\pm$0.0015}}&0.002\text{\tiny{$\pm$0.0003}}&0.055&0.602\text{\tiny{$\pm$0.0022}}&0.532\text{\tiny{$\pm$0.0025}}&0.07\\ \hline
-3&0.014\text{\tiny{$\pm$0.0008}}&0.001\text{\tiny{$\pm$0.0002}}&0.013&0.601\text{\tiny{$\pm$0.0033}}&0.583\text{\tiny{$\pm$0.0036}}&0.018\\ \hline
-4&0.015\text{\tiny{$\pm$0.0025}}&0.002\text{\tiny{$\pm$0.0003}}&0.013&0.607\text{\tiny{$\pm$0.0027}}&0.592\text{\tiny{$\pm$0.0031}}&0.015\\ \hline
-5&0.003\text{\tiny{$\pm$0.0005}}&0.005\text{\tiny{$\pm$0.0006}}&-0.002&0.604\text{\tiny{$\pm$0.0032}}&0.605\text{\tiny{$\pm$0.0029}}&-0.001\\ \hline

\end{tabular}
\caption{Explaining layers for ResNet: \emph{ClaDec} has larger reconstruction loss but the evaluation classifier on reconstructions from \emph{ClaDec} achieves higher accuracy.} \label{tab:layRes}
\vspace{-6pt}
\end{table*}

 Quantitative Results in Tables \ref{tab:lay} and \ref{tab:layRes} contain two key messages: First, the reconstruction loss is lower for \emph{RefAE} than for \emph{ClaDec}. This follows since \emph{RefAE} is optimized entirely towards minimal reconstruction loss of the original inputs. Second, the classification (evaluation) accuracy is (almost always) higher, when training the evaluation classifier $C_{Eval}$ using reconstructions from \emph{ClaDec} rather than from \emph{RefAE}. This behavior is not obvious and it might be seen as surprising since the reconstructions from \emph{ClaDec} are poorer according to the reconstruction loss. That is, they contain less information about the original input than those from \emph{RefAE}. However, it seems that the ``right'' information is encoded using a better suited representation. Only for the very last layer, i.e., the linear layer, the accuracy for the evaluation classifier tends to be higher if trained on reconstructions from \emph{RefAE}. The last layer acts most discriminatively and has lower dimensions, i.e., for the classifier often only relatively few neurons are active in the last layer. The reconstructions from \emph{ClaDec} are often very different from the original and of poor quality, i.e., blurry, or resembling fairly different looking, prototypical objects (though of the same category). For CIFAR-100 also once the reconstruction loss is lower for \emph{ClaDec} though not by a large margin and a visual assessment generally attests lower quality to reconstructions. Aiming for a smooth, average-like reconstruction seems to be beneficial.
 
 Aside from these two key observations there are a set of other noteworthy behaviors: The reconstruction loss increases the more encoder layers, i.e., the more the input is transformed. In absolute numbers, the impact is significantly stronger for \emph{ClaDec}. The difference between \emph{RefAE} and \emph{ClaDec} increases the closer the layer to explain is to the output.
 For Resnet (Table \ref{tab:layRes}) the accuarcy of the evaluation classifier is fairly constant across layers for \emph{ClaDec} (except for the very last layer, where it drops) and the reconstruction loss is significantly lower for both \emph{RefAE} and \emph{ClaDec} compared to VGG. This is not unexpected, since Resnets do not downsample that much (up to the last layer the spatial dimension is still 4x4) and it contains residual connections that facilitate information flow. However, also for ResNets the reconstruction error clearly increases for \emph{ClaDec} with more layers.  \\



\begin{figure}
\centering \includegraphics[width=\linewidth]{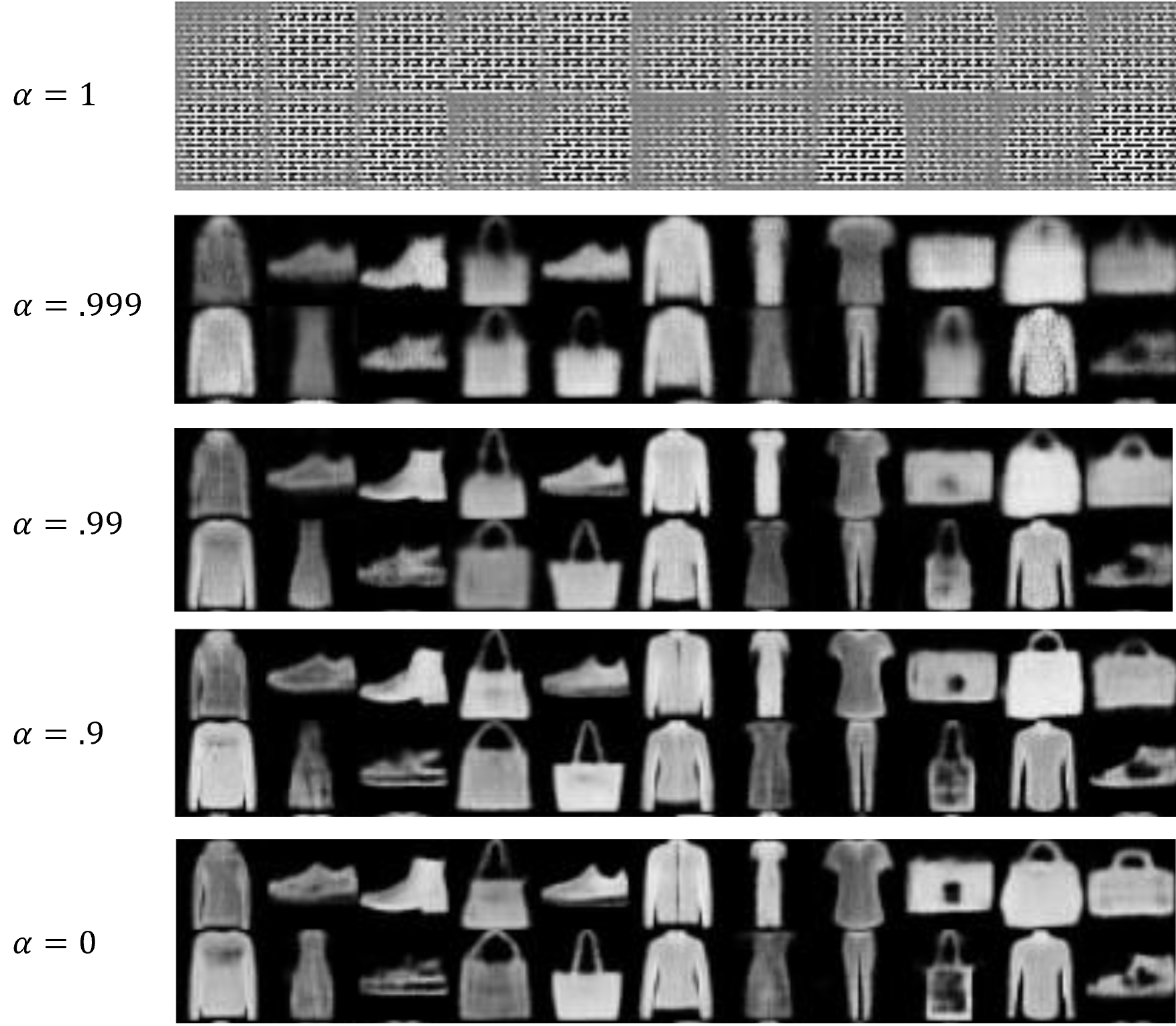}
  \caption{Adding classification loss ($\alpha>0$) yields worse reconstructions for the last conv. layer. Using classification loss only ($\alpha=1$), reconstructions are not human recognizable.} \label{fig:trade}
\end{figure}

\begin{table}
\footnotesize
\begin{tabular}{|l| l| l|l| l|}\hline
$\alpha$&Total Loss  \emph{ClaDec} &Rec Loss& Classifier Loss &Acc $C^{ClaDec}_{Eval}$  \\ \hline
1.0&0.01\text{\tiny{$\pm$0.003}}&285.5\text{\tiny{$\pm$52.01}}&0.0\text{\tiny{$\pm$0.0}}&0.903\text{\tiny{$\pm$0.003}}\\ \hline
0.999&0.03\text{\tiny{$\pm$0.002}}&25.4\text{\tiny{$\pm$0.929}}&0.03\text{\tiny{$\pm$0.0}}&0.903\text{\tiny{$\pm$0.002}}\\ \hline
0.9&0.84\text{\tiny{$\pm$0.013}}&8.35\text{\tiny{$\pm$0.12}}&0.75\text{\tiny{$\pm$0.011}}&0.901\text{\tiny{$\pm$0.002}}\\ \hline
0&7.49\text{\tiny{$\pm$0.112}}&7.49\text{\tiny{$\pm$0.112}}&4.4\text{\tiny{$\pm$0.254}}&0.882\text{\tiny{$\pm$0.004}}\\ \hline
 \end{tabular}
\caption{For VGG, adding classification loss $\alpha>0$ (Equation \ref{eq:alpha}) yields worse reconstructions, but higher evaluation accuracy}\label{tab:acc}
\end{table}

\begin{table*}
\footnotesize
\begin{tabular}{|l| l| l|l| l|}\hline
$\alpha$&Total Loss \emph{ClaDec} &Rec Loss& Classifier Loss &Acc $C^{ClaDec}_{Eval}$  \\ \hline
1.0&0.009\text{\tiny{$\pm$0.001}}&416.5\text{\tiny{$\pm$76.69}}&0.0\text{\tiny{$\pm$0.0}}&0.912\text{\tiny{$\pm$0.003}}\\ \hline
0.999&0.043\text{\tiny{$\pm$0.003}}&34.5\text{\tiny{$\pm$1.6}}&0.035\text{\tiny{$\pm$0.002}}&0.911\text{\tiny{$\pm$0.002}}\\ \hline
0.99&0.195\text{\tiny{$\pm$0.009}}&19.0\text{\tiny{$\pm$0.704}}&0.188\text{\tiny{$\pm$0.007}}&0.911\text{\tiny{$\pm$0.003}}\\ \hline
0.9&1.33\text{\tiny{$\pm$0.017}}&13.3\text{\tiny{$\pm$0.164}}&1.193\text{\tiny{$\pm$0.015}}&0.909\text{\tiny{$\pm$0.002}}\\ \hline
0&12.2\text{\tiny{$\pm$0.196}}&12.2\text{\tiny{$\pm$0.196}}&4.936\text{\tiny{$\pm$0.057}}&0.898\text{\tiny{$\pm$0.005}}\\ \hline
 \end{tabular}
\caption{For ResNet, adding classification loss $\alpha>0$  yields worse reconstructions, but higher evaluation accuracy}\label{tab:accRes} 
\vspace{-12pt}
\end{table*}

\subsection{Fidelity and Comprehensibility Trade-off} \label{sec:fidel}
We assess the impact of adding a classification loss (Figure \ref{fig:deco}) to modulate using a parameter $\alpha \in [0,1]$ how much the classifier model impacts reconstructions. Neglecting reconstruction loss, i.e., $\alpha=1$, yields non-interpretable reconstructions as shown in the first row in Figure \ref{fig:trade}. Already modest reconstruction loss leads to well-human recognizable shapes. The quality of reconstructions in terms of sharpness and amount of captured detail constantly improves the more emphasis is put on reconstruction loss. It also becomes evident that the neural network learns ``prototypical'' samples (or features) towards which reconstructed samples are being optimized. For example, the shape of handbag handles shows much more diversity for values of $\alpha$ close to 0, it is fairly uniform for relatively large values of $\alpha$. 
Thus, the parameter $\alpha$ provides a means to reconstruct a compromise between the sample that yields minimal classification loss and a sample that is true to the input. It suggests that areas of the reconstruction of \emph{ClaDec} that are similar to the original input are also similar to a ``prototype'' that minimizes classification loss. The network can recognize them well, whereas areas that are strongly modified, resemble parts that seem non-aligned with ``the prototype'' encoded in the network.

Quantitative results in Tables \ref{tab:acc} and \ref{tab:accRes} shows that evaluation accuracy increases when adding a classification loss, i.e., $\alpha>0$ yields an accuracy above 90\% whereas $\alpha=0$ gives about 88\%. Reconstructions that are stronger influenced by the model to explain (larger $\alpha$) are more truthful to the model, but they exhibit larger differences from the original inputs. Choosing $\alpha$ slightly above the minimum, i.e., larger than 0, already has a strong impact.

\subsection{Impact of Encoder Training} \label{sec:expAE}
\begin{figure}
  \centering
  \includegraphics[width=\linewidth]{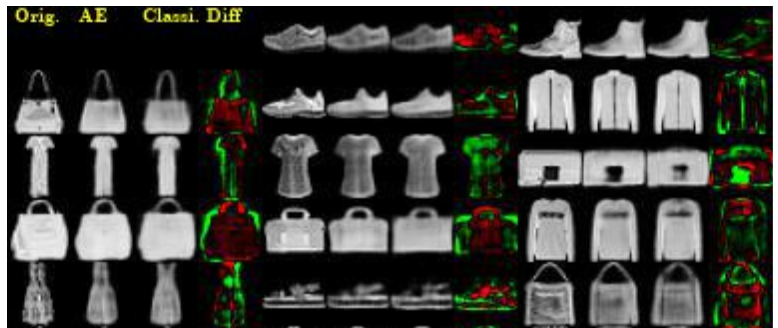}
  \caption{Comparison of original inputs and reconstructions using the last conv. layer of the encoder in Table  \ref{tab:arch} without any training of the classifier in \emph{ClaDec}. Green indicating brighter values of \emph{ClaDec} than \emph{RefAE} and red the opposite} \label{fig:showNoTrain}
\end{figure}

Comparing reconstructions of the \emph{RefAE} with those of \emph{ClaDec} for an untrained classifier might be used to assess the relevance of training the encoder of the classifier or an AE itself, i.e., ``How does training an encoder impact reconstructions compared to a non-trained encoder?''.
Figure \ref{fig:showNoTrain} shows reconstructions if the classifier is not trained at all. The figure suggests that a trained encoder does lead to encodings that allow to better reconstruct original inputs than an untrained encoder utilizing randomly initialized layers.  While sharpness is generally comparable for both reconstructions, there are several samples where shapes of objects are altered, or details differ. Overall, this behavior can be understood using theory on random projections as we explain next for our quantitative results in Table \ref{tab:ep}.
The table shows for \emph{ClaDec} that classifiers that are trained longer and, therefore, achieve higher validation accuracy also lead to better accuracy for the evaluation classifier. More surprising is the dependence of the reconstruction loss on the number of training epochs. For \emph{ClaDec}, it is lowest without any training, increases quickly and then steadily decreases again. This pattern is highly statistically significant. We conducted t-tests to verify that means between subsequent rows are different, yielding p-values below 0.01. The fact that an untrained network, i.e., using random weights, achieves lower reconstruction loss than the trained classifier can be explained as follows: First, it should be noted that the reconstruction loss using random weights is significantly higher than for the reference architecture, where the encoder is optimized. Second, it is known from extreme learning, e.g., \cite{sun2017}, that encoders with randomly chosen weights can yield good results, if just the decoder is optimized. More generally, good encoding properties might be deducted from the behavior of random projections formulated in the Johnson–Lindenstrauss lemma. It says that distances are well-preserved in a low-dimensional space originating from random projections. The theorem is commonly used in machine learning, e.g., \cite{sch13,sch14}. Training the classifier seems to destroy some of the desirable properties of random initialization by focusing on information needed for classification (but not for reconstruction) -- as motivated theoretically. The reconstruction improves with more training, indicating that the initial encodings are noisy. 
\begin{table*}
\begin{tabular}{|l|l||l|l|l|| l|l|l|}\hline
Train. Epochs  &Acc. of  & Rec Loss & Rec Loss &$\Delta$Rec&Acc Eval &Acc Eval &$\Delta$Acc  \\
of Classifier   &Classifier & \emph{ClaDec} & \emph{RefAE}& &\emph{ClaDec} & \emph{RefAE}&  \\
to be expl.   & to be expl.& & &  & &&  \\
\hline
0 & 0.1\tiny{$\pm0.0$}&5.445\tiny{$\pm0.164$}&3.333\tiny{$\pm0.04$}&2.112&0.85\tiny{$\pm0.003$}&0.886\tiny{$\pm0.004$}&-0.036 \\ \hline
1 & 0.506\tiny{$\pm0.012$}&6.417\tiny{$\pm0.152$}&3.3\tiny{$\pm0.038$}&3.118&0.88\tiny{$\pm0.002$}&0.888\tiny{$\pm0.003$}&-0.007 \\ \hline
4 & 0.885\tiny{$\pm0.003$}&6.608\tiny{$\pm0.079$}&3.299\tiny{$\pm0.049$}&3.309&0.893\tiny{$\pm0.004$}&0.893\tiny{$\pm0.004$}&0.0 \\ \hline
16 & 0.902\tiny{$\pm0.003$}&6.233\tiny{$\pm0.145$}&3.334\tiny{$\pm0.062$}&2.898&0.896\tiny{$\pm0.005$}&0.891\tiny{$\pm0.003$}&0.005 \\ \hline
64 & 0.904\tiny{$\pm0.003$}&6.081\tiny{$\pm0.069$}&3.341\tiny{$\pm0.062$}&2.74&0.895\tiny{$\pm0.002$}&0.889\tiny{$\pm0.004$}&0.006 \\ \hline
\end{tabular}
\caption{Impact of Classifier Accuracy (modulated through training epochs): Evaluation Accuracy increases with higher classifier accuracy, behavior of rec.loss follows an inverted U shape.}\label{tab:ep}
\vspace{-16pt}
\end{table*}

\subsection{Explaining Subsets of Layer Activations} \label{sec:layAct} 
\begin{figure}
  \centering
  \includegraphics[width=\linewidth]{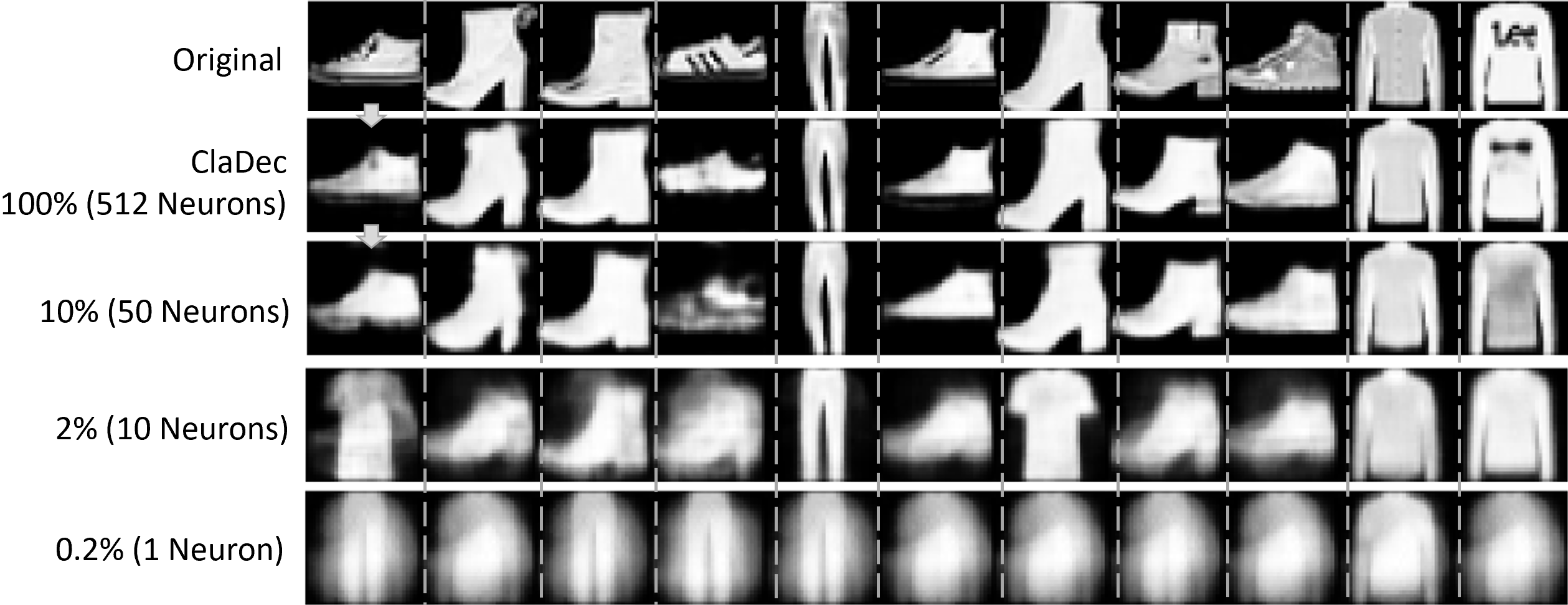}
  \caption{Reconstructions for \emph{ClaDec} when using subsets of all activations of varying size for the last conv layer of VGG. Reconstructions quality decreases with fewer neurons.} \label{fig:subsetChangeSize}
\end{figure}

\begin{figure}
  \centering
  \includegraphics[width=\linewidth]{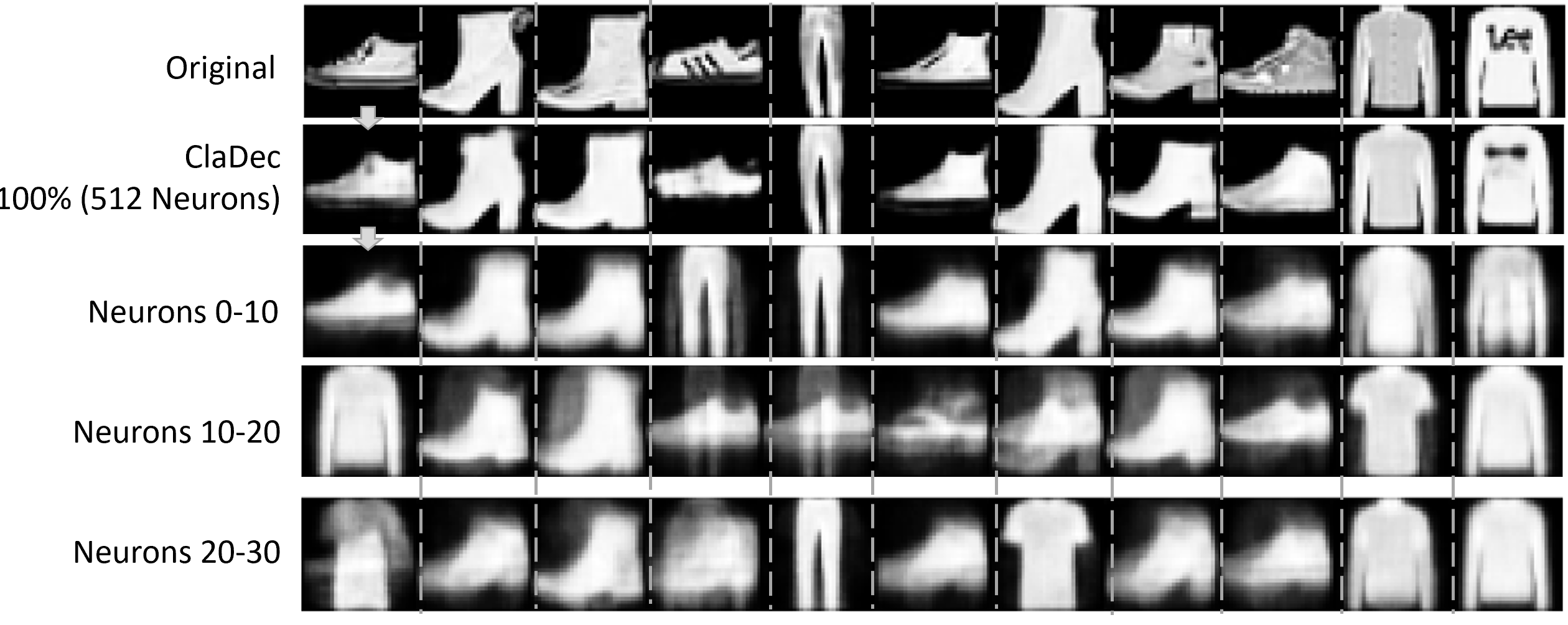}
  \caption{Reconstructions for \emph{ClaDec} when using disjoint subsets of all activations of the last conv layer of VGG. Class associations of subsets are well-recognizable.} \label{fig:subsetDisjoin}
\end{figure}

We aim to assess reconstructions that are only based on some neurons of a layer. While neurons in lower layers often encode generic, class-independent information, neurons in higher layers are more strongly associated with specific concepts found in a dataset or a specific class. That is, neurons from upper layers often show large activations for one class or few classes only. This leads to the hypothesis that a subset of neurons might describe some classes well and others not so well. Since activations of different neurons often correlate, the idea that a neuron encodes one specific pattern that can only be reconstructed using this specific neuron is generally not valid \cite{fong18}. However, more information on a concept certainly helps in obtaining a better reconstruction as we shall show.\\
We train \emph{ClaDec} using only a subset of neurons, i.e., their activations. We pick a subset $S \subseteq L$ where $L$ are all neurons of a layer. We reconstruct the input only based on the activations of this subset $S$ of neurons. In our evaluation, we decode subsets of activations of the last convolutional layer. In Figure \ref{fig:subsetChangeSize} we vary the subset size so that subsets with fewer neurons are subsets of those with more neurons.  Formally,  let $L'$ and $L''$ be two subsets of $L$, i.e., $L',L''\subset L$. If $|L'|<|L''|$ then $L' \subset L''$. In Figure \ref{fig:subsetDisjoin} we maintain the same subset size but use disjoint subsets, that is $L'\cap L'' =\{\}$. Neurons can be chosen arbitrarily. We chose subsets consisting of a sequence of neurons, which is as good as choosing them randomly since indexes of neurons and their parameters and semantics bear no relationship.\\

Overall, it is well visible that using fewer activations to reconstruct leads to less detailed and more blurry reconstructions - see Figures \ref{fig:subsetDisjoin} and \ref{fig:subsetChangeSize}. However, it is well-known that deep learning networks contain significant redundancy and inter-dependencies \cite{fong18}. Therefore, using only a fraction of all neurons is likely to lead to samples that are still resembling the input to a significant degree, i.e., neuron activations tend to correlate.\\
When using disjoint sets (Figure \ref{fig:subsetDisjoin}) some samples are very poorly reconstructed compared to those based on all neurons. Poor reconstructions happen when we use only a small number of neurons of a layer. These neurons relate to concepts that are mostly associated with one class, i.e., we use 2\% of all neurons of the last convolutional layer. Thus, for some classes $A$, there are no neurons that contain any strongly related concept. That is, the decoder cannot distinguish inputs, i.e., subsets of activations, belonging to such classes $A$ well. That is, to the decoder samples from classes $A$ look alike. In this case, reconstructions resemble overlays corresponding to prototypes (or averages) of multiple classes. They appear in low intensity and blurry. More precisely, the (weak) overlays show the classes the decoder is uncertain about. For example, the figure shows overlays of sneakers and pants in row 4 (Neurons 10-20). Thus, blurriness indicates a lack of information (in the activations) to reconstruct input details or even just the class. It also becomes apparent that in case there are neurons associated with a class, reconstructions are fairly decent, in the sense that they resemble a well-recognizable class prototype. However, a closer look reveals that they lack many details compared to reconstructions based on all neurons. \\
When using smaller subsets of neurons (Figure \ref{fig:subsetChangeSize}) one can see a steady decrease in quality. For a single neuron classes can only be guessed. For a particular class and a subset of neurons $S$ reconstructions could be good because most neurons in $S$ contribute some information to the reconstruction or a few neurons in $S$ are very strongly related to the class and others are not at all.

\subsection{Comparing \emph{ClaDec} Explanations to Learnt Class Prototypes}\label{sec:proto}

\begin{figure}
  \centering
  \includegraphics[width=\linewidth]{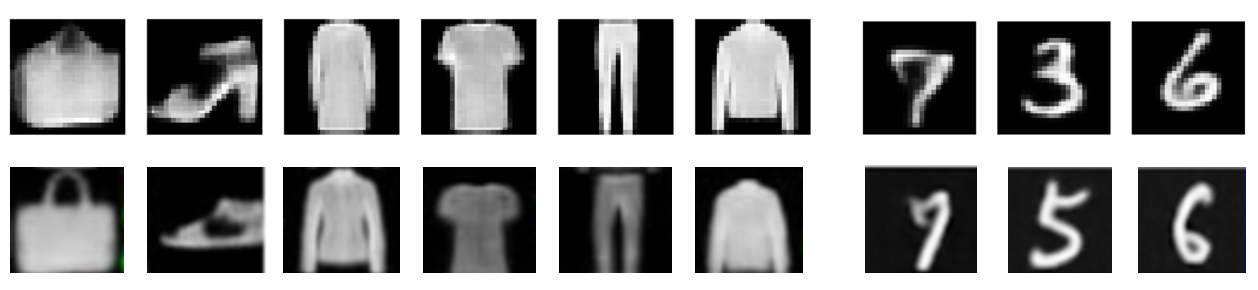}
  \caption{Comparison of explanations from \emph{ClaDec} from last conv layer (bottom row) with class prototypes from \cite{li18de} (top row) for MNIST and FashionMNIST} \label{fig:compProto}
\end{figure}

Figure \ref{fig:compProto} shows a subset of the 15 class prototypes learnt in \cite{li18de}.  \cite{li18de} proposes a special architecture to learn prototypes that are also used during the classification process. The number of prototypes to be learnt is a hyperparameter set to 15 for the datasets in \cite{li18de}. \emph{ClaDec} can be said to compute a distinct reconstruction for each of the (thousands of) samples that might appear prototypical. Figure \ref{fig:compProto} indicates that explanations from \emph{ClaDec} and prototypes from \cite{li18de} both focus on information relevant for classification, e.g., they do not show texture and stripes. The class prototypes from \cite{li18de} do not appear to be a ``typical'' (or average) representation but bear similarity to an explanation of \emph{ClaDec} for a seemingly randomly chosen sample. That is, class prototypes contain concepts that are not widely present across samples. For instance, the ``7'' contains several small uncommon artifacts, and the handbag is also a specific type of bag with a handle that is clearly shorter than those of the average.\footnote{We inspected about 100 handbags from the dataset to derive this conclusion.}  In the proposed architecture \cite{li18de} these class prototypes reside in the ``prototype'' layer that is followed by a fully connected layer as the last layer before the output is computed (using a Softmax). This implies that the class prototypes are also encoding concepts that occur at this abstraction level. As the prototypes are not reconstructed based on the very last layer but are followed by a fully connected layer, classification depends on the combination of multiple prototypes. For instance, there might be a prototype showing a red T-shirt and a black pullover. When a red pullover is classified as a pullover, the match with both prototypes might contribute to the decision of the pullover (though the red T-shirt prototype might be stronger associated with the T-shirt class). As such a prototype showing a sample of a particular class might lack concepts that are also common for this class.\\

To summarize, the explanations of both methods lead to similar insights about the general behavior of the classifier. While \cite{li18de} has explainability built into the model, \emph{ClaDec} also comes with a set of advantages: It is model-agnostic (i.e., to all state-of-the-art models), it explains individual decisions (i.e., it highlights for each sample what is relevant and what not), it allows to explain different layers and subsets thereof. 

\subsection{Impact of Occlusions of Inputs} \label{sec:occ} 
We qualitatively and quantitatively assess the impact of input occlusions. The idea is to assess the change in appearance of reconstructions in our qualitative evaluation and the change in prediction accuracy when altering more or less relevant parts of the input. The idea to obstruct parts of the input to understand their relevance is common in XAI, e.g., \cite{pet18}. For saliency maps obtaining relevance is trivial since saliency maps output a relevance score for each pixel. Thus, to get the relevance of an arbitrary area, relevance score of pixels can be summed. For reconstructions from \emph{ClaDec} determining the relevance of a pixel and area is more tricky since it is based on reconstructing concepts and not on computing scores for (input) pixels. To obtain a pixel relevance score, we use the idea that if parts relevant to the classification in the input are distorted, the classifier's layer activation also gets heavily distorted. Thus, reconstructions by ClaDec get strongly distorted. If parts of the input get occluded that are not impacting the classifier, then the activations and, consequently, the reconstructions $\hat{X}_E$ by \emph{ClaDec} should not change much. Therefore, we can assess how much the overall reconstructed image gets altered due to occluding parts of the input. Thus, assume we occlude a set of pixels $O$, then the relevance $R(O)$ of these pixels is given by the sum of differences in reconstructions between the reconstruction $\hat{X}_{E}$ for the original input $X$ by \emph{ClaDec} and the reconstruction $\hat{X}^{occ}_{E}$ of the partially occluded input $X^{occ}$ by \emph{ClaDec}. Formally, we define
$$R^{ClaDec}(O):=\sum_{i \in O} ||\hat{X}^{occ}_{E,i}-\hat{X}_{E,i}||^2$$
In our evaluation, we consider for each input 16 options for occlusion. Each occlusion having the shape of a square of size $12 \times 12$ pixels with the upper left corner $(x_u,y_u)$ given by $x_u, y_u \in [0,6,12,20]$. We replace the occluded pixels with the mean value of all pixels in the training data, i.e., a gray tune.

In our quantitative evaluation, we compute the accuracy of the classifier $C$ to explain occluded inputs. We compare the accuracy when occluding least and most relevant parts of the input according to \emph{ClaDec}, i.e., the measure $R^{ClaDec}(O,X)$, and according to GradCAM, where we defined the relevance $R^{GradCAM}(O,X)$ as being the sum of the value of the saliency map $S(X)$ normalized to $[0,1]$ of pixel $i \in O$, i.e., $R^{GradCAM}(O,X):=\sum_{i \in O} S_i(X)$. In particular, we are interested in each image in the occlusion out of the 16 possibilities yielding minimum (or maximum) relevance according to an explainability method, i.e., either GradCAM or \emph{ClaDec}. For a relevance score $R \in \{R^{ClaDec},R^{GradCAM}\}$, we define $X_{O,min}$ with an occlusion $O_{min}$ of areas of minimum relevance. Since there can be many occlusions with the same minimum relevance, $O_{min}$ is chosen uniformly at random out of all of the occlusions $\{O' |  R(O',X)=\min_{O} R(O,X)\}$ yielding minimum relevance. $X_{O,max}$ and $O_{max}$ are defined analogously. We define $Acc(C(X_{O,min}))$ as the accuracy of the classifier $C$ to explain on samples $X_{O,min}$, i.e. samples $X$ where an area of minimum relevance is occluded. 

\begin{figure}
  \centering
  \includegraphics[width=\linewidth]{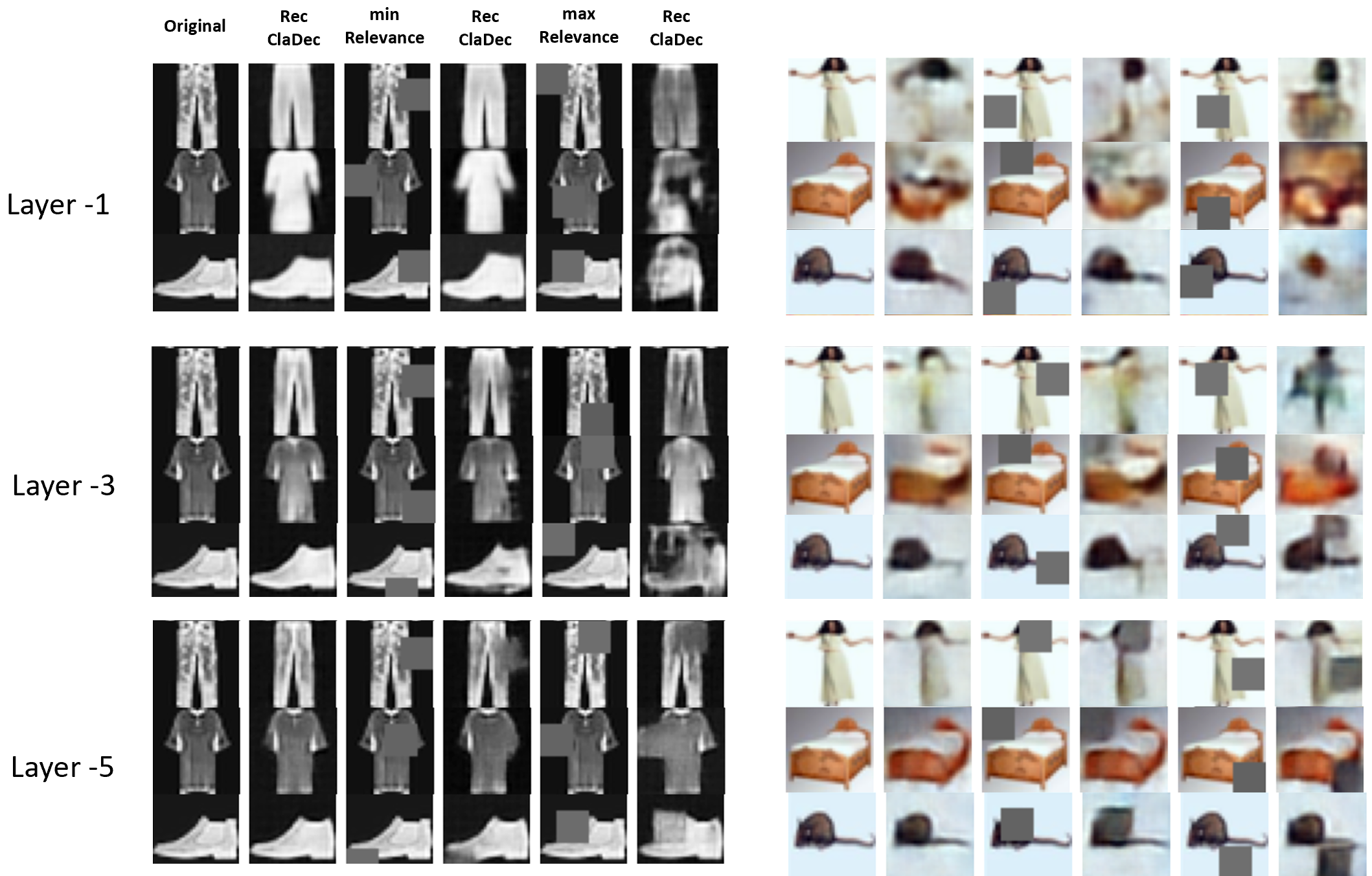}
  \caption{Reconstructions for \emph{ClaDec} when occluding parts of the image. Columns are original image $X$, its reconstruction, occluded image $X_{O,min}$, its reconstruction, $X_{O,max}$ and its reconstruction for FasionMNIST and CIFAR-100 for VGG} \label{fig:occ}
\end{figure}

\noindent \textbf{Qualitative Evaluation: } Figure \ref{fig:occ} shows the most and least relevant occlusions. It can be seen that occluding areas in inputs deemed more relevant lead to larger distortion in the reconstructions. Relevance is not stable across layers, i.e., the areas with minimum (and maximum) relevance scores depend on the layer. Occluding an area can lead to specific features at a layer to activate or fail to activate. Convolutional kernels have a larger receptive field and more semantic features for higher-level layers. This implies that activating a feature at a higher level might still be possible, although parts of the input that activate it are occluded.
Furthermore, the impact on the reconstruction is expected to cover a larger area and yield more extreme distortions in the reconstruction when a higher-level feature is not activating. This can be well-seen in Figure \ref{fig:occ} when comparing the reconstruction of occluding the maximum relevance area for the shoe (left-hand side) or bed (right-hand side) layer -1 and layer -5. For layer -5 the reconstructed image is almost identical to the original except for the occluded square, whereas for layer -1 it looks completely different. Aside from that, lower layers allow reconstructing the image better than higher layers since fewer transformations (and possibly loss due to downsampling) of input information has occurred. The gray area indicating an occlusion is well-visible in the reconstruction in lower layers in Figure \ref{fig:occ}, and the rest of the object is also commonly undistorted. Since our relevance measure is based on the difference between the reconstructed image without occlusion and with occlusion, the occlusion that maximizes relevance is related primarily to the difference between the pixels of the occluded area in the two compared reconstructions since the rest is almost identical, i.e., it has low relevance. This can lead to the situation where the maximum relevance area is part of an (uninformative) background that seems irrelevant for classification and the minimum relevance area contains the object to be classified. This is illustrated well in Figure \ref{fig:occ} for the mouse and the woman on layer -5. The area of maximum relevance is the white background in both cases since the difference between white and gray pixels is very large.
In contrast, it is the object itself for minimum relevance since it is darker and therefore, the difference (and relevance) is smaller. Lower layers of the classifier do not alter the information flow very much, i.e., they do not filter much information of the input, no matter where the occlusions occur. Thus, if only a few layers of the classifier processed the occluded input, it can be well-reconstructed independent of the occlusion location. However, if more layers are involved, the reconstruction depends heavily on the activation of the input features that are semantically meaningful and tight to specific input areas.

Reconstructions also differ strongly depending on whether an area of minimum or maximum relevance is distorted. As expected, the visual appearance does not vary much across layers when areas of minimum relevance are occluded since all images contain areas that do not impact layer representations of upper layers, i.e., they do not help discriminate between classes. The object to be recognized remains mostly intact in the reconstruction. For maximum relevance, this is not the case when reconstructions from activations of upper layers are computed. The reconstructed image can appear differently because the occluded area and semantics can change. This becomes most apparent when considering the outcome of the quantitative evaluation.

\noindent \textbf{Quantitative Evaluation: }Tables \ref{tab:occ} and \ref{tab:occ2} show the impact on accuracy on the classifier to explain when areas of minimum and maximum relevance are occluded based on GradCAM and \emph{ClaDec} for various networks and datasets. Aligned with our qualitative evaluation, for \emph{ClaDec} we can observe that occluding areas of maximum relevance have a larger impact on accuracy than areas of lower relevance. Furthermore, the impact tends to be larger for upper layers than for lower layers for areas of maximum relevance. For ResNet (Table \ref{tab:occ2}) the differences in accuracy ($\Delta$ Acc) can be small for lower layers or even negative. This is aligned with our qualitative analysis, indicating that reconstructions are not depending strongly on the occluded area as well as the observation that the relevance measure is not so much tied to the object to be classified, e.g., the background rather than the object to be classified might yield maximum relevance. While the absolute numbers vary across networks and datasets, the general behavior tends to be very similar. Most notably, the difference in ($\Delta$ Acc) varies less across layers for MNIST than CIFAR-100, and the impact is also largest for MNIST. MNIST is a fairly simple dataset, where only a relatively small area of the input decides on the class, e.g., adding a small line can change the class from 6 to 8 or from 3 to 9. Thus, occlusions can be chosen to have no impact or to change the class with a high likelihood.
Interestingly, while Resnets perform better in general for MNIST they are more sensitive to occlusions than VGG networks, e.g., accuracy is lower for an occluded area. We attribute this to the fact that Resnets are downsampling differently. They maintain more spatial information in early layers and also up to the last dense layer and use average-pooling, whereas VGG networks downsample more aggressively already in lower layers by max-pooling. This allows occluded areas that might resemble  features similar to actual classes to directly influence the output. For example a gray square indicating occlusions in a background area might appear as a 0 or parts of an 8 or 9. 

\noindent \emph{Comparing GradCAM and ClaDec: } For GradCAM, only lower layers yield a significant difference in accuracy when comparing occluded areas of minimum and maximum relevance for VGG but not so for Resnet. This is a consequence of the spatial extent of the layer. If GradCAM is applied to layers with little spatial extend, relevance scores are imprecise. In the most extreme case, they are identical for all pixels, i.e., if the layer has a spatial extend of 1x1, the upsampling of relevance scores to the full 32x32 image yields equal scores across pixels. As already described, Resnet preserves spatial dimensions up to higher layers than VGG. This explains the different behavior for GradCAM shown in Tables \ref{tab:occ} and \ref{tab:occ2}. For Resnet, removing maximum relevant areas yields larger changes in classifier performance except for MNIST. However, for MNIST and layer -1 we observed very large standard deviations when removing the minimum relevant areas using occlusions. Most interesting is arguably the question of which of the two explainability methods yields better outcomes. This might be judged based on whether occluding more or less relevant parts strongly impacts accuracy, i.e., $\Delta$ Acc. Choosing areas randomly as minimum and maximum relveant would yield a $\Delta$ Acc of zero. Thus, larger $\Delta$ Acc means that a method can better anticipate what is relevant and irrelevant for a classifier.  Overall, $\Delta$ Acc is larger for \emph{ClaDec}, in particular for VGG. This is remarkable, in particular, since the relevance measure used in the computation is well-suited for GradCAM but rather poor for \emph{ClaDec}. GradCAM is developed to output relevance score for pixels which can be aggregated to get the relevance of an (occluded) area. \emph{ClaDec} requires a rather complex computation of relevance score using differences between reconstructions which introduces additional noise, in particular for datasets such as CIFAR-100, where reconstructions are generally not of very high quality. Noise, i.e., randomly choosing areas as minimum and maximum relevant, reduces $\Delta$ Acc. Therefore, it is somewhat surprising that  overall our evaluation indicates that \emph{ClaDec} can better identify relevant areas than \emph{GradCAM} that is developed for this purpose. However, we believe that both methods are of high value, particularly for datasets where reconstructions of \emph{ClaDec} are not of very high quality.

\begin{table*}
\begin{tabular}{|l|l|l||l|l|l|l|}\hline
Layer  &    \multicolumn{3}{|c|}{\emph{ClaDec}} & \multicolumn{3}{|c|}{\emph{GradCAM} } \\ \cline{2-7}
& \multicolumn{2}{|c|}{Accuracy} & $\Delta Acc$ &   \multicolumn{2}{|c|}{Accuracy} & $\Delta Acc$
  \\ \cline{2-3} \cline{5-6}
  & $C(X_{O,max})$  & $C(X_{O,min})$ &  &  $C(X_{O,max})$ &  $C(X_{O,min})$  & 
  \\ \hline
\multicolumn{7}{|c|}{CIFAR-100} \\ \hline
-1 &   0.277\tiny{\text{$\pm$}0.006} & 0.399\tiny{\text{$\pm$}0.005} & 0.122 & 0.334\tiny{\text{$\pm$}0.004} & 0.334\tiny{\text{$\pm$}0.004} & 0.0\\
-3 &    0.298\tiny{\text{$\pm$}0.006} & 0.386\tiny{\text{$\pm$}0.003} & 0.088 & 0.355\tiny{\text{$\pm$}0.003} & 0.377\tiny{\text{$\pm$}0.002} & 0.023 \\
-5 &    0.325\tiny{\text{$\pm$}0.008} & 0.367\tiny{\text{$\pm$}0.002} & 0.042 & 0.321\tiny{\text{$\pm$}0.002} & 0.371\tiny{\text{$\pm$}0.003} & 0.05 \\  \hline
\multicolumn{7}{|c|}{FashionMNIST} \\ \hline
-1 &   0.783\tiny{\text{$\pm$}0.017} & 0.907\tiny{\text{$\pm$}0.001} & 0.124 & 0.859\tiny{\text{$\pm$}0.006} & 0.859\tiny{\text{$\pm$}0.006} & 0.0 \\
-3 &    0.771\tiny{\text{$\pm$}0.018} & 0.906\tiny{\text{$\pm$}0.002} & 0.135 & 0.878\tiny{\text{$\pm$}0.002} & 0.882\tiny{\text{$\pm$}0.003} & 0.004\\
-5 &  0.824\tiny{\text{$\pm$}0.013} & 0.888\tiny{\text{$\pm$}0.004} & 0.064 & 0.843\tiny{\text{$\pm$}0.011} & 0.879\tiny{\text{$\pm$}0.005} & 0.036  \\  \hline
\multicolumn{7}{|c|}{MNIST} \\ \hline
-1 & 0.772\tiny{\text{$\pm$}0.028} & 0.993\tiny{\text{$\pm$}0.0} & 0.221 & 0.914\tiny{\text{$\pm$}0.006} & 0.914\tiny{\text{$\pm$}0.006} & 0.0 \\
-3 &   0.76\tiny{\text{$\pm$}0.011} & 0.994\tiny{\text{$\pm$}0.001} & 0.234 & 0.989\tiny{\text{$\pm$}0.0} & 0.988\tiny{\text{$\pm$}0.001} & 0.0 \\
-5 &  0.766\tiny{\text{$\pm$}0.011} & 0.994\tiny{\text{$\pm$}0.001} & 0.227 & 0.858\tiny{\text{$\pm$}0.008} & 0.947\tiny{\text{$\pm$}0.007} & 0.089  \\
\hline
\end{tabular}
\caption{Accuracy when occluding least and most relevant parts of input for \emph{ClaDec} and \emph{GradCAM} for VGG}\label{tab:occ}
\vspace{-16pt}
\end{table*}

\begin{table*}
\begin{tabular}{|l|l|l||l|l|l|l|}\hline
Layer  &    \multicolumn{3}{|c|}{\emph{ClaDec}} & \multicolumn{3}{|c|}{\emph{GradCAM} } \\ \cline{2-7}
& \multicolumn{2}{|c|}{Accuracy} & $\Delta Acc$ &   \multicolumn{2}{|c|}{Accuracy} & $\Delta Acc$
  \\ \cline{2-3} \cline{5-6}
  & $C(X_{O,max})$  & $C(X_{O,min})$ &  &  $C(X_{O,max})$ &  $C(X_{O,min})$  &  
  \\ \hline
\multicolumn{7}{|c|}{CIFAR-100} \\ \hline
-1 &  0.377\tiny{\text{$\pm$}0.007} & 0.52\tiny{\text{$\pm$}0.004} & 0.143 & 0.381\tiny{\text{$\pm$}0.006} & 0.533\tiny{\text{$\pm$}0.004} & 0.152 \\
-3 &   0.458\tiny{\text{$\pm$}0.005} & 0.462\tiny{\text{$\pm$}0.005} & 0.004 & 0.398\tiny{\text{$\pm$}0.003} & 0.502\tiny{\text{$\pm$}0.003} & 0.105 \\
-5 &    0.466\tiny{\text{$\pm$}0.005} & 0.453\tiny{\text{$\pm$}0.006} & -0.013 & 0.441\tiny{\text{$\pm$}0.002} & 0.456\tiny{\text{$\pm$}0.005} & 0.015 \\  \hline
\multicolumn{7}{|c|}{FashionMNIST} \\ \hline
-1 &    0.814\tiny{\text{$\pm$}0.023} & 0.914\tiny{\text{$\pm$}0.003} & 0.1 & 0.875\tiny{\text{$\pm$}0.006} & 0.912\tiny{\text{$\pm$}0.007} & 0.037 \\
-3 &   0.876\tiny{\text{$\pm$}0.012} & 0.887\tiny{\text{$\pm$}0.005} & 0.011 & 0.874\tiny{\text{$\pm$}0.004} & 0.893\tiny{\text{$\pm$}0.01} & 0.019 \\
-5 &   0.878\tiny{\text{$\pm$}0.006} & 0.885\tiny{\text{$\pm$}0.008} & 0.006 & 0.888\tiny{\text{$\pm$}0.004} & 0.889\tiny{\text{$\pm$}0.004} & 0.001 \\  \hline
\multicolumn{7}{|c|}{MNIST} \\ \hline
-1 &   0.548\tiny{\text{$\pm$}0.147} & 0.877\tiny{\text{$\pm$}0.075} & 0.329 & 0.757\tiny{\text{$\pm$}0.093} & 0.774\tiny{\text{$\pm$}0.165} & 0.023\\
-3 &   0.573\tiny{\text{$\pm$}0.043} & 0.824\tiny{\text{$\pm$}0.075} & 0.251 & 0.523\tiny{\text{$\pm$}0.066} & 0.809\tiny{\text{$\pm$}0.046} & 0.286 \\
-5 &   0.509\tiny{\text{$\pm$}0.064} & 0.773\tiny{\text{$\pm$}0.072} & 0.264 & 0.564\tiny{\text{$\pm$}0.109} & 0.566\tiny{\text{$\pm$}0.051} & 0.001 \\
\hline
\end{tabular}
\caption{Accuracy when occluding least and most relevant parts of input for \emph{ClaDec} and \emph{GradCAM} for Resnet}\label{tab:occ2}
\vspace{-16pt}
\end{table*}

\subsection{User Study} \label{sec:huExp} 
We conducted an experiment asking humans for their judgment. Our goals were to  (i) assess whether non-experts can make sense of our explanations, (ii) assess whether humans can identify a diverse set of concepts in the explanations, i.e., if they can determine many concepts abstracted by the AI such as texture, shape, color, etc., and (iii) conduct an exploratory analysis of free-text responses in a qualitative manner to highlight other interesting findings. We recruited people from the general public who were mostly unfamiliar with convolutional neural networks and deep learning.  That is, they were neither aware of the concept of layers nor of the workings of an autoencoder. We decided to focus on a simple scenario where the reconstructions from \emph{RefAE} are similar to the original image. We neglected differences that could be attributed due to distortion of the decoder since they were mostly relatively small compared to differences between the reconstructions from \emph{ClaDec} and we did not want to overload inexperienced participants in a single experiment. 

\begin{figure}
  \centering
  \includegraphics[width=\linewidth]{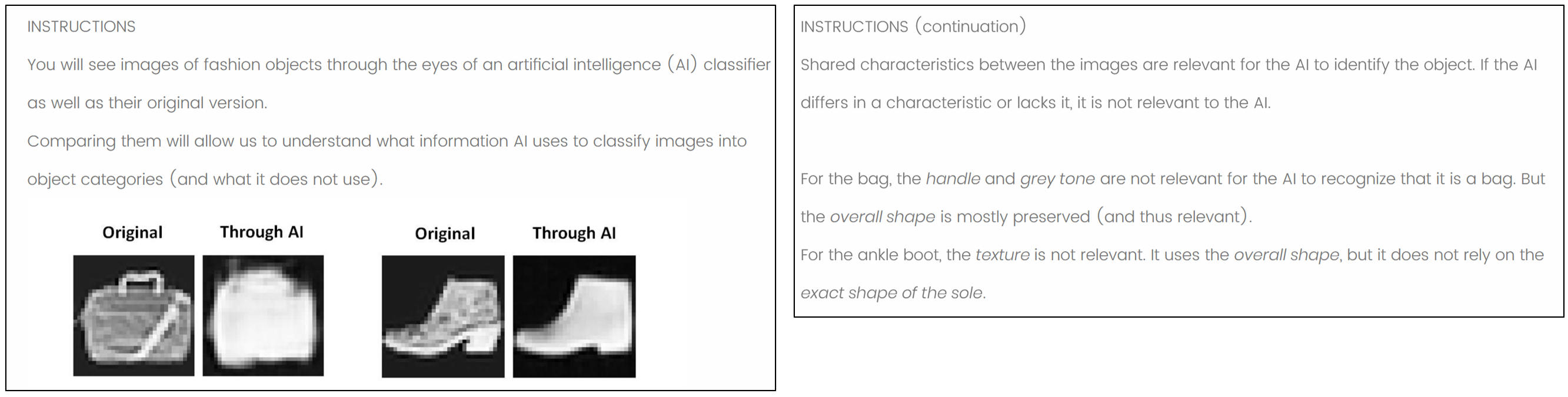}
  \caption{Introduction for participants of human study} \label{fig:studyintro}
\end{figure}

We used the second-to-last layer of the VGG-11 network and 100 samples of the Fashion-MNIST dataset. That is, a sample consists of the original image as well as the reconstruction from \emph{ClaDec}. 
We recruited 59 participants through the platform ``Prolific''. Their demographics and prior knowledge are summarized in Table \ref{tab:user}. 

\begin{table}
\scriptsize
\begin{tabular}{lllll}

\emph{Question}                                              & \multicolumn{2}{l}{\emph{Response   Distribution} }                             &                          &               \\ \hline
Gender                                                & 30 males                      & \multicolumn{2}{l}{29 females}                                     &               \\
Age                                                   & 41 (age 18-24)          & 12 (age 35-44)                     & \multicolumn{2}{l}{6 (age 35-44)} \\
Education                                             & 26 (some college)             & 13 (high school)                         & 14 (4 years degree)     & 6 (Other)     \\
Do you know what ...           &&&&\\
... deep  learning is?            & 49 (no) & 10 (yes) & &  \\
... a CNN is? & 54 (no) & 5 Yes  &                  &              
\end{tabular}
 	\caption{Participants' demographics and prior knowledge}  \label{tab:user} 
 	\vspace{-6pt}
\end{table}

\begin{figure}
  \centering
  \includegraphics[width=0.7\linewidth]{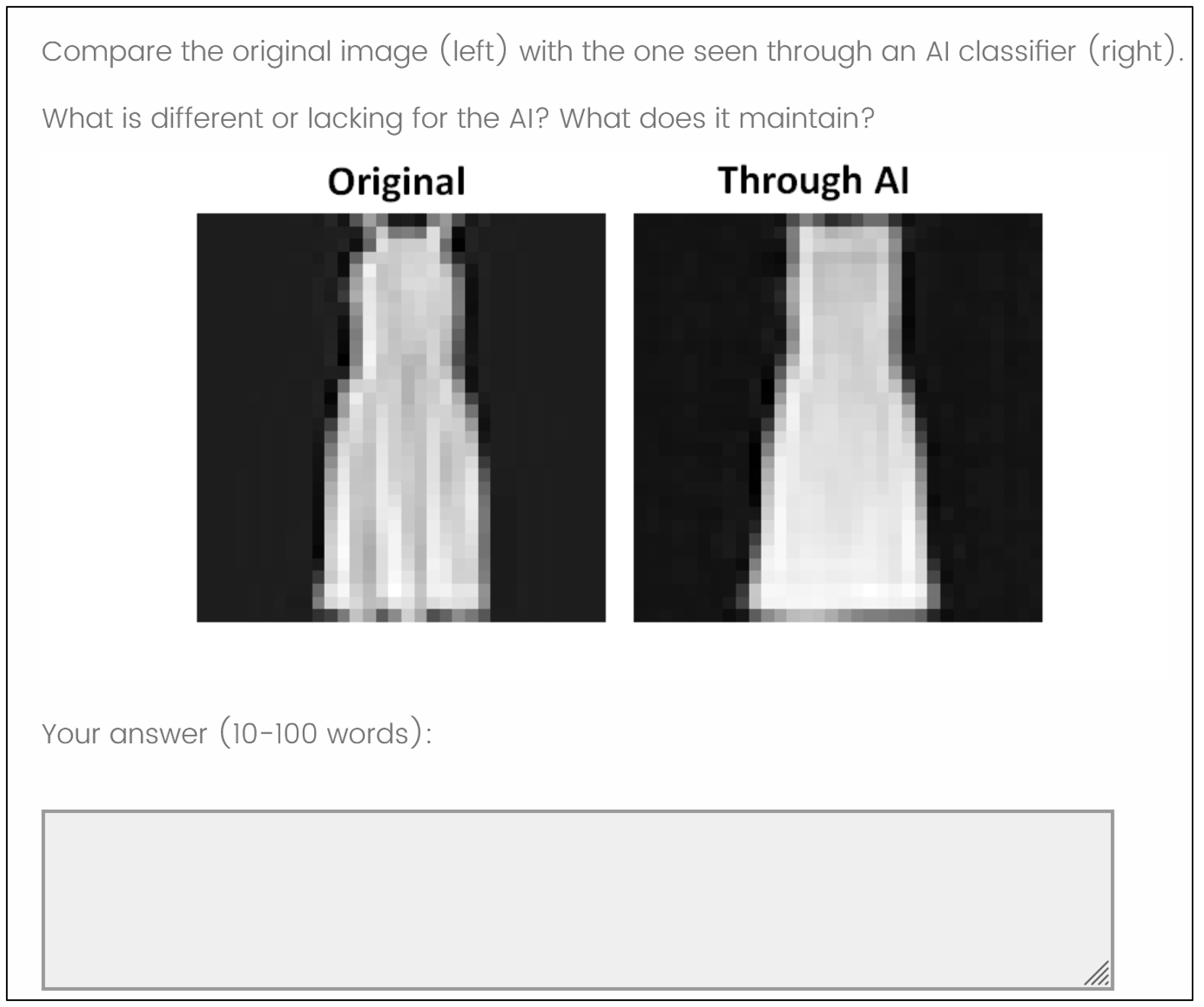}
  \caption{Question presented to participant} \label{fig:studyQue}
\end{figure}

Participants were given a short general introduction on AI and image recognition and a more detailed one for the task shown in Figure \ref{fig:studyintro}. Participants were asked to compare the original image and the one from \emph{ClaDec}, i.e., the image as seen through an AI classifier as shown in the example in Figure \ref{fig:studyQue}. Each participant analyzed five randomly selected image pairs qualitatively using textual analysis so that each pair was analyzed equivalently often. \\
\textbf{Analysis and Findings:}
We obtained 295 free-text answers (5 from each of the 59 participants), e.g., most images had three responses and a few just two.
We did not exclude any answers. To understand whether participants reached valid conclusions, we checked for  erroneous  judgments by participants. We investigated whether participants claimed images were (almost) identical although there were clear differences.  We found that about 1\%, i.e., three responses, claimed to see no differences though we (and other participants) could identify clear differences. We found that about 1\% of responses contained objectively false claims, e.g., they claimed that the original and reconstructed objects differed (significantly) in size though they did not. Many responses contained imprecise language, i.e., respondents used generic terms like ``design differs'', which could imply differences in shape, texture, color, etc. Furthermore, terms like texture, patterns and characteristics were often used interchangeably. Overall, 2\% of responses are incorrect, i.e., they contradict responses of other participants and the ground truth (defined by the author team). The remaining 98\% of responses are deemed valid, i.e., they describe well-visible differences. However, some of the responses contain linguistic imprecision, and most responses only contain a subset of all visible differences. \\

We identified generic concepts, i.e., comparison criteria  that can be applied to multiple classes through a qualitative data analysis method, i.e., ``open coding''\cite{cor90}. We found that responses used as criteria for comparison: ``outer'' or ``overall'' shape (66\% of replies contained this concept), a feature or specific characteristic (40\% of replies), color or grey tone (29\%), pattern or texture (24\%), logo or brand name (15\%), shading (10\%), design (8\%), sharpness or blurriness (5\%),  perspective/orientation of object (3\%), artifacts introduced by AI (1\%), contrast (of images) (1\%), image depth (1\%), symmetry/skewness (1\%). Overall, the responses of users show that our explainability process of comparing the original image and the reconstructed image by \emph{ClaDec} allows even non-experts to identify a rich set of concepts. Note that some concepts were rare, e.g., only a few explanations contained obvious artifacts introduced by the reconstruction or a change in perspective. Also, our introduction mentioned the concepts ``overall shape'', ``texture'' and ``grey tone'', which might have primed users to use these more often.\\

Most concepts were used primarily to indicate differences except for (overall) shape, which was mostly mentioned to be preserved (more than 90\% of responses). Color and pattern/textures were mentioned commonly in contexts where they differed and where they were identical. Thus, the comparison seems to be biased towards differences. We also found that about 15\% of answers only focused on differences though we explicitly asked also what the AI maintains. Emphasizing differences rather than commonalities is not unexpected since our visual system is primed to focus attention on movement and change, i.e., differences \cite{fran03}.\\ 

Other interesting findings were that a few responses (about 2\%) stated that the reconstructions tend to look like another class due to the absence of specific features or other changes. For example, one response was ``AI preserves the shape, but erases the patterns and also the zipper, so this image is no longer a sweater but a shirt''.  Furthermore, while most replies tended to indicate that there were either little differences between the original and the reconstruction, or the reconstruction lost some information of the original, some replies also indicated the opposite, i.e., that the reconstructions are ``better'' or easier to recognize than the original. In the most extreme case, particpants could recognize the object reconstructed by the AI, but not the original (1\% of replies), e.g., one reply was ``I can't tell what the original image shows, but through AI I can easily tell that it's a dress''. This is not unexpected since the AI tends to reconstruct prototypical images replacing uncommon features with more common ones. When it comes to responses, aligned with our perception, users mostly mentioned reconstructions to be more blurry but interestingly, a few responses also indicated the opposite (1\%), i.e., for two images where the pattern contained many dark patches making the contour more difficult to recognize.\\

In summary, our user study shows that non-experts can analyze our explanations and identify a rich set of concepts within images that are relevant (or irrelevant) to the classification process. To interpret the explanations, another small step is needed: If the concepts in the input differ from concepts in the ClaDec reconstruction (but not from those of the RefAE), the input concept differs significantly from those encoded by the classifier, i.e., the sample differs significantly from a prototypical instance and details of the input concept that could not be reconstructed are not relevant to distinguish between classes. 

If the input shape differs from the shape of the ClaDec reconstruction (but not from the shape of the RefAE), this means that the classifier does not need to be able to reconstruct it well to classify the object. The sample differs significantly from a prototypical instance.

\section{Discussion and Future Work}
To create explanations, \emph{ClaDec} must be trained on a dataset that should be similar to the training dataset though no labels are needed. This is a disadvantage compared to other methods such as GradCAM that can be used in situ. If the dataset is too small, the reference AE and \emph{ClaDec} both cannot produce high quality reconstructions limiting the value of our method. This is often manifested in poor reconstructions of the \emph{RefAE}, which correlates strongly with difficult to comprehend reconstructions $\hat{X}_E$ by the \emph{ClaDec}. In our case, the CIFAR-100 dataset is the most complex and the qualitative analysis suggests that only coarse differences between reconstructions from \emph{RefAE} and \emph{ClaDec} such as shapes and color tunes are easy to identify by humans. The CIFAR-100 classifiers also perform poorest among all datasets. Thus, it can be concluded that explanations are better for well-performing classifiers that are typically also based on a large dataset. We might use similar tricks used for classifiers to improve our explainability method, i.e. the decoders. For example, data augmentation techniques could be used to enhance a dataset. That is, even if the classifier is trained on non-augmented data, we might train \emph{RefAE} and \emph{ClaDec} using data augmentation. Furthermore, pre-trained AE architectures could be leveraged using transfer learning. Our evaluation supports the idea that reconstruction-based techniques should employ a reference since generative models tend to introduce some distortion and artifacts. Non-reconstruction based explanation methods do not require comparing to a reference. However, for a sufficiently large dataset and an adequate  architecture that yields reconstructions of the \emph{RefAE} that are not distinguishable from the original, this step is also not necessary. For the MNIST dataset, which is relatively large compared to the number of classes and the complexity of samples, reconstructions are very good for both \emph{RefAE} and they are essentially only needed to highlight differences of the topmost layer, where the number of dimensions is small, and thus distortion are largest. On the contrary, for a more complex dataset with many more classes but the same number of samples like CIFAR-100, the comparison to the \emph{RefAE} is helpful. While our method still provides many interesting insights, interpretation is more tricky due to distortions in reconstruction. However, other methods like GradCAM are of little help in such cases, i.e., when the difference between the spatial extent of the feature map and the reconstructed input is very large. This is also a finding of our occlusion-based analysis. Overall, we believe that reconstruction-based methods and saliency-based methods are complementary.\\
Our user study confirmed that users could easily identify differences in actual input images and reconstructions by ClaDec. It might also be interesting to perform a quantitative study, e.g., asking users explicitly if they notice differences with respect to a specific concept. Furthermore, our user study focused on the case where \emph{RefAE} and original inputs show little differences. This limitation could also be remedied with another user study.\\

Our work also touches on fundamental questions in deep learning, i.e., the information bottleneck \cite{saxe19,gei20}. Our work contributes to this discussion since it indicates that information is discarded from layer to layer because reconstructions get poorer for upper layers as shown qualitatively and quantitatively. Still, more work is needed in this direction.

Another aspect is computation time for an explanation. Our method incurs computational setup costs due to the initial training of \emph{ClaDec} and \emph{RefAE} as well as for computing each explanation. Roughly speaking, the costs for explaining are about the same as for model training. Some methods like GradCAM do not require any setup costs, while methods like LIME require training of many approximate models, which can incur much higher computational costs. 


\section{Conclusions} 
Our explanation method synthesizes human understandable inputs based on layer activations or subsets thereof. It takes into account distortions originating from the reconstruction process. Rather than pinpointing to individual neurons or parts of an input, we are interested in understanding what information of the input can be reconstructed. We believe that our method might form the basis for many more methods that further expand and contribute to the field of explainability. %


\section{Declarations}
\subsection{Availability of data and material }
Code is at \url{https://github.com/JohnTailor/ClaDec}. All data used is public.

\subsection{Competing interests}
The authors declare that they have no 
competing interests. 
\subsection{Ethics Approval}
Not applicable
\subsection{Consent to participate}
Not applicable
\subsection{Consent for publication}
Not applicable

\subsection{Funding}
Not Applicable
\subsection{Authors Contribution}
MV: Proof reading, GradCAM visualizations
JS: all the rest
\subsection{Acknowledgements}
We thank Jeroen van Doorenmalen for valuable discussions.

None

\bibliographystyle{spmpsci.bst}
\bibliography{refs}

\begin{thebibliography}{10}
\providecommand{\url}[1]{{#1}}
\providecommand{\urlprefix}{URL }
\expandafter\ifx\csname urlstyle\endcsname\relax
  \providecommand{\doi}[1]{DOI~\discretionary{}{}{}#1}\else
  \providecommand{\doi}{DOI~\discretionary{}{}{}\begingroup
  \urlstyle{rm}\Url}\fi

\bibitem{adadi2018peeking}
Adadi, A., Berrada, M.: Peeking inside the black-box: A survey on explainable
  artificial intelligence (xai).
\newblock IEEE Access \textbf{6} (2018)

\bibitem{adebayo2018sanity}
Adebayo, J., Gilmer, J., Muelly, M., Goodfellow, I., Hardt, M., Kim, B.: Sanity
  checks for saliency maps.
\newblock In: Neural Information Processing Systems (2018)

\bibitem{aga20}
Agarwal, C., Nguyen, A.: Explaining image classifiers by removing input
  features using generative models.
\newblock In: Proceedings of the Asian Conference on Computer Vision (2020)

\bibitem{bach2015pixel}
Bach, S., Binder, A., Montavon, G., Klauschen, F., M{\"u}ller, K.R., Samek, W.:
  On pixel-wise explanations for non-linear classifier decisions by layer-wise
  relevance propagation.
\newblock PloS one \textbf{10}(7) (2015)

\bibitem{bal89}
Baldi, P., Hornik, K.: Neural networks and principal component analysis:
  Learning from examples without local minima.
\newblock Neural networks \textbf{2}(1), 53--58 (1989)

\bibitem{bar20}
Barbalau, A., Cosma, A., Ionescu, R.T., Popescu, M.: A generic and
  model-agnostic exemplar synthetization framework for explainable ai.
\newblock In: ECML-PKDD (2020)

\bibitem{che18}
Chen, C., Li, O., Tao, D., Barnett, A., Rudin, C., Su, J.K.: This looks like
  that: Deep learning for interpretable image recognition.
\newblock Advances in Neural Information Processing Systems  (2019)

\bibitem{conf21}
Confalonieri, R., Coba, L., Wagner, B., Besold, T.R.: A historical perspective
  of explainable artificial intelligence.
\newblock Wiley Interdisciplinary Reviews: Data Mining and Knowledge Discovery
  \textbf{11}(1) (2021)

\bibitem{conf19}
Confalonieri, R., Weyde, T., Besold, T.R., Moscoso~del Prado~Mart{\'\i}n, F.:
  Trepan reloaded: A knowledge-driven approach to explaining black-box models.
\newblock In: ECAI 2020 (2020)

\bibitem{cor90}
Corbin, J.M., Strauss, A.: Grounded theory research: Procedures, canons, and
  evaluative criteria.
\newblock Qualitative sociology \textbf{13}(1), 3--21 (1990)

\bibitem{dec93}
Deco, G., Finnoff, W., Zimmermann, H.: Elimination of overtraining by a mutual
  information network.
\newblock In: Int. Conference on Artificial Neural Networks (1993)

\bibitem{van20}
van Doorenmalen, J., Menkovski, V.: Evaluation of cnn performance in
  semantically relevant latent spaces.
\newblock In: Int. Symposium on Intelligent Data Analysis (2020)

\bibitem{du16}
Du, B., Xiong, W., Wu, J., Zhang, L., Zhang, L., Tao, D.: Stacked convolutional
  denoising auto-encoders for feature representation.
\newblock IEEE transactions on cybernetics \textbf{47}(4), 1017--1027 (2016)

\bibitem{fong18}
Fong, R., Vedaldi, A.: Net2vec: Quantifying and explaining how concepts are
  encoded by filters in deep neural networks.
\newblock In: Proceedings of the IEEE conference on computer vision and pattern
  recognition, pp. 8730--8738 (2018)

\bibitem{fran03}
Franconeri, S.L., Simons, D.J.: Moving and looming stimuli capture attention.
\newblock Perception \& psychophysics \textbf{65}(7), 999--1010 (2003)

\bibitem{gei20}
Geiger, B.C.: On information plane analyses of neural network classifiers--a
  review.
\newblock IEEE Transactions on Neural Networks and Learning Systems  (2021)

\bibitem{gho19}
Ghorbani, A., Abid, A., Zou, J.: Interpretation of neural networks is fragile.
\newblock In: AAAI Conference on Artificial Intelligence (2019)

\bibitem{ghor19}
Ghorbani, A., Wexler, J., Zou, J.Y., Kim, B.: Towards automatic concept-based
  explanations.
\newblock In: Advances in Neural Information Processing Systems (2019)

\bibitem{goo16}
Goodfellow, I., Bengio, Y., Courville, A.: Deep learning.
\newblock MIT press (2016)

\bibitem{gui19bl}
Guidotti, R., Monreale, A., Matwin, S., Pedreschi, D.: Black box explanation by
  learning image exemplars in the latent feature space.
\newblock In: Joint European Conference on Machine Learning and Knowledge
  Discovery in Databases, pp. 189--205 (2019)

\bibitem{kim17}
Kim, B., Wattenberg, M., Gilmer, J., Cai, C., Wexler, J., Viegas, F., et~al.:
  Interpretability beyond feature attribution: Quantitative testing with
  concept activation vectors (tcav).
\newblock In: International conference on machine learning (2018)

\bibitem{kin19}
Kindermans, P.J., Hooker, S., Adebayo, J., Alber, M., Sch{\"u}tt, K.T.,
  D{\"a}hne, S., Erhan, D., Kim, B.: The (un) reliability of saliency methods.
\newblock In: Explainable AI: Interpreting, Explaining and Visualizing Deep
  Learning (2019)

\bibitem{koh2017}
Koh, P.W., Liang, P.: Understanding black-box predictions via influence
  functions.
\newblock In: Proc. of Int. Conference on Machine Learning (2017)

\bibitem{li18de}
Li, O., Liu, H., Chen, C., Rudin, C.: Deep learning for case-based reasoning
  through prototypes: A neural network that explains its predictions.
\newblock In: Proceedings of the AAAI Conference on Artificial Intelligence
  (2018)

\bibitem{liu19tow}
Liu, W., Li, R., Zheng, M., Karanam, S., Wu, Z., Bhanu, B., Radke, R.J., Camps,
  O.: Towards visually explaining variational autoencoders.
\newblock In: Proceedings of the IEEE/CVF Conference on Computer Vision and
  Pattern Recognition (2020)

\bibitem{ngu16}
Nguyen, A., Dosovitskiy, A., Yosinski, J., Brox, T., Clune, J.: Synthesizing
  the preferred inputs for neurons in neural networks via deep generator
  networks.
\newblock In: Advances in neural information processing systems, pp. 3387--3395
  (2016)

\bibitem{pet18}
Petsiuk, V., Das, A., Saenko, K.: Rise: Randomized input sampling for
  explanation of black-box models.
\newblock arXiv preprint arXiv:1806.07421  (2018)

\bibitem{qi19co}
Qi, F., Lin, C., Shi, G., Li, H.: A convolutional encoder-decoder network with
  skip connections for saliency prediction.
\newblock IEEE Access \textbf{7}, 60428--60438 (2019)

\bibitem{raf20}
Rafegas, I., Vanrell, M., Alexandre, L.A., Arias, G.: Understanding trained
  cnns by indexing neuron selectivity.
\newblock Pattern Recognition Letters \textbf{136}, 318--325 (2020)

\bibitem{ribeiro2016should}
Ribeiro, M.T., Singh, S., Guestrin, C.: Why should i trust you?: Explaining the
  predictions of any classifier.
\newblock In: SIGKDD (2016)

\bibitem{rud19}
Rudin, C.: Stop explaining black box machine learning models for high stakes
  decisions and use interpretable models instead.
\newblock Nature Machine Intelligence \textbf{1}(5) (2019)

\bibitem{saxe19}
Saxe, A.M., Bansal, Y., Dapello, J., Advani, M., Kolchinsky, A., Tracey, B.D.,
  Cox, D.D.: On the information bottleneck theory of deep learning.
\newblock Journal of Statistical Mechanics: Theory and Experiment
  \textbf{2019}(12), 124020 (2019)

\bibitem{sch20hu}
Schneider, J.: Human-to-{AI} coach: Improving human inputs to {AI} systems.
\newblock In: International Symposium on Intelligent Data Analysis (2020)

\bibitem{schneider2019pers}
Schneider, J., Handali, J.P.: Personalized explanation for machine learning: a
  conceptualization.
\newblock In: European Conference on Information Systems (ECIS) (2019)

\bibitem{schne22d}
Schneider, J., Meske, C., Vlachos, M.: Deceptive {AI} explanations: Creation
  and detection.
\newblock In: International Conference on Agents and Artificial Intelligence
  (ICAART) (2022)

\bibitem{sch13}
Schneider, J., Vlachos, M.: Fast parameterless density-based clustering via
  random projections.
\newblock In: Proc. of the international conference on Information \& Knowledge
  Management(CIKM) (2013)

\bibitem{sch14}
Schneider, J., Vlachos, M.: On randomly projected hierarchical clustering with
  guarantees.
\newblock In: Proceedings of the SIAM International Conference on Data Mining,
  pp. 407--415 (2014)

\bibitem{sch20ref}
Schneider, J., Vlachos, M.: Reflective-net: Learning from explanations.
\newblock In: arxiv: 2011.13986 (2020)

\bibitem{selvaraju2017grad}
Selvaraju, R.R., Cogswell, M., Das, A., Vedantam, R., Parikh, D., Batra, D.:
  Grad-cam: Visual explanations from deep networks via gradient-based
  localization.
\newblock In: IEEE International Conference on Computer Vision (ICCV), pp.
  618--626 (2017)

\bibitem{shr17}
Shrikumar, A., Greenside, P., Kundaje, A.: Learning important features through
  propagating activation differences.
\newblock In: Int. Conf. on Machine Learning (2017)

\bibitem{sim13}
Simonyan, K., Vedaldi, A., Zisserman, A.: Deep inside convolutional networks:
  Visualising image classification models and saliency maps.
\newblock In: In Workshop at International Conference on Learning
  Representations (2014)

\bibitem{sun2017}
Sun, K., Zhang, J., Zhang, C., Hu, J.: Generalized extreme learning machine
  autoencoder and a new deep neural network.
\newblock Neurocomputing \textbf{230}, 374--381 (2017)

\bibitem{vin10}
Vincent, P., Larochelle, H., Lajoie, I., Bengio, Y., Manzagol, P.A.: Stacked
  denoising autoencoders: Learning useful representations in a deep network
  with a local denoising criterion.
\newblock Journal of machine learning research \textbf{11}(Dec), 3371--3408
  (2010)

\bibitem{wu2020exp}
Wu, W., Su, Y., Chen, X., Zhao, S., King, I., Lyu, M.R., Tai, Y.W.: Towards
  global explanations of convolutional neural networks with concept
  attribution.
\newblock In: Proceedings of the IEEE/CVF Conference on Computer Vision and
  Pattern Recognition, pp. 8652--8661 (2020)

\bibitem{yang19}
Yang, F., Du, M., Hu, X.: Evaluating explanation without ground truth in
  interpretable machine learning.
\newblock arXiv preprint arXiv:1907.06831  (2019)

\bibitem{yeh19}
Yeh, C.K., Hsieh, C.Y., Suggala, A., Inouye, D.I., Ravikumar, P.K.: On the (in)
  fidelity and sensitivity of explanations.
\newblock In: Advances in Neural Information Processing Systems, pp.
  10965--10976 (2019)

\bibitem{yos15}
Yosinski, J., Clune, J., Fuchs, T., Lipson, H.: Understanding neural networks
  through deep visualization.
\newblock In: In ICML Workshop on Deep Learning (2015)

\bibitem{zeil14}
Zeiler, M.D., Fergus, R.: Visualizing and understanding convolutional networks.
\newblock In: European conference on computer vision (2014)

\end{thebibliography}

\end{document}